\documentclass[11pt]{article}
\usepackage[preprint]{acl}
\usepackage{times}
\usepackage{latexsym}
\usepackage[T1]{fontenc}
\usepackage[utf8]{inputenc}
\usepackage{microtype}
\usepackage{inconsolata}
\usepackage{graphicx}
\usepackage{xcolor}
\usepackage{amsmath,amssymb}
\usepackage{booktabs,tabularx,array}
\usepackage{placeins}
\graphicspath{{figures/}}
\usepackage{xurl}
\newcommand{\ED}{\mathrm{ED}}
\newcommand{\RRI}{\mathrm{RRI}}
\newcommand{\WEP}{\mathrm{WEP}_{30}}
\newcommand{\DCE}{\mathrm{DCE}}
\newcommand{\FPA}{\textsc{FPA}}
\newcommand{\AMEM}{A-MEM\textsuperscript{1}}
\newcommand{\MEMZERO}{Mem0\textsuperscript{2}}
\newcommand{\COMM}{CoM\textsuperscript{3}}
\newcommand{\LANGMEM}{LangMem\textsuperscript{4}}
\newcommand{\memorysources}{\par\vspace{1pt}\begin{minipage}{\columnwidth}\raggedright\footnotesize \textsuperscript{1} A-MEM~\citep{ref2}; \textsuperscript{2} Mem0~\citep{ref3}; \textsuperscript{3} CoM~\citep{ref20}; \textsuperscript{4} LangMem~\citep{ref21}.\end{minipage}}

\title{Memory Echo in Single-Agent Long-Term Memory:\\Lifecycle Evaluation and Provenance-Guided Mitigation}
\author{
            \textbf{Yufeng Li\textsuperscript{1}\thanks{%
    Equal contribution. Yufeng Li: \protect\nolinkurl{yufeng03@stu.ecnu.edu.cn}; Sheng Yao: \protect\nolinkurl{yaosheng@usc.edu}. \textsuperscript{\textdagger}Corresponding author: Peng Zeng. Email: \protect\nolinkurl{pzeng@sei.ecnu.edu.cn}.
  },
  Sheng Yao\textsuperscript{2}\footnotemark[1],
  Peng Zeng\textsuperscript{1,\textdagger},
  Dezhang Kong\textsuperscript{3}} \\
  \textbf{Shi Lin\textsuperscript{4},
  Changting Lin\textsuperscript{3},
  Xun Wang\textsuperscript{4},
  Meng Han\textsuperscript{3}} \\[4pt]
  \textsuperscript{1}East China Normal University, Shanghai, China \\
  \textsuperscript{2}University of Southern California, Los Angeles, USA \\
  \textsuperscript{3}Zhejiang University, Hangzhou, China \\
  \textsuperscript{4}Zhejiang Gongshang University, Hangzhou, China
}

\begin{document}
\maketitle

\begin{abstract}
Long-term memory supports agents across sessions. Repeated retrieval and write-back can let one historical claim occupy multiple retrieval positions and influence current answers without independent evidence. Existing work examines retrieval, updating, and question answering, leaving same-source record accumulation and post-correction persistence underexplored. Motivated by research in social cognition and memory psychology on repeated exposure, retrieval, and revision, we define \emph{Memory Echo} as the excess retrieval exposure gained by historical information from the same source without additional independent evidence, and introduce MEchoBench to evaluate its emergence, correction resistance, persistence, and effect on answers. To address this phenomenon, we introduce the Family--Provenance--Authority framework (FPA). Specifically, FPA selects one representative per source family, conditions the eligibility of old records for the answer context on verified corrections, and constrains final answers with authoritative evidence. In paired replay on 190 fixed questions across four memory frameworks, FPA raises accuracy from 0.825 to 0.920 with DeepSeek V4 Flash and from 0.782 to 0.916 with Kimi K2.6; mean Decision Contamination Effect falls from 0.106 to 0.030 and from 0.109 to 0.038, respectively. These results show that agent memory can preserve traceable history without allowing repeated historical information to govern current decisions. The code is available at \nolinkurl{https://github.com/OrcasSam/memory-echo-benchmark}.
\end{abstract}

\section{Introduction}\label{sec:intro}
\begin{figure}[t]
\centering
\includegraphics[width=0.49\textwidth]{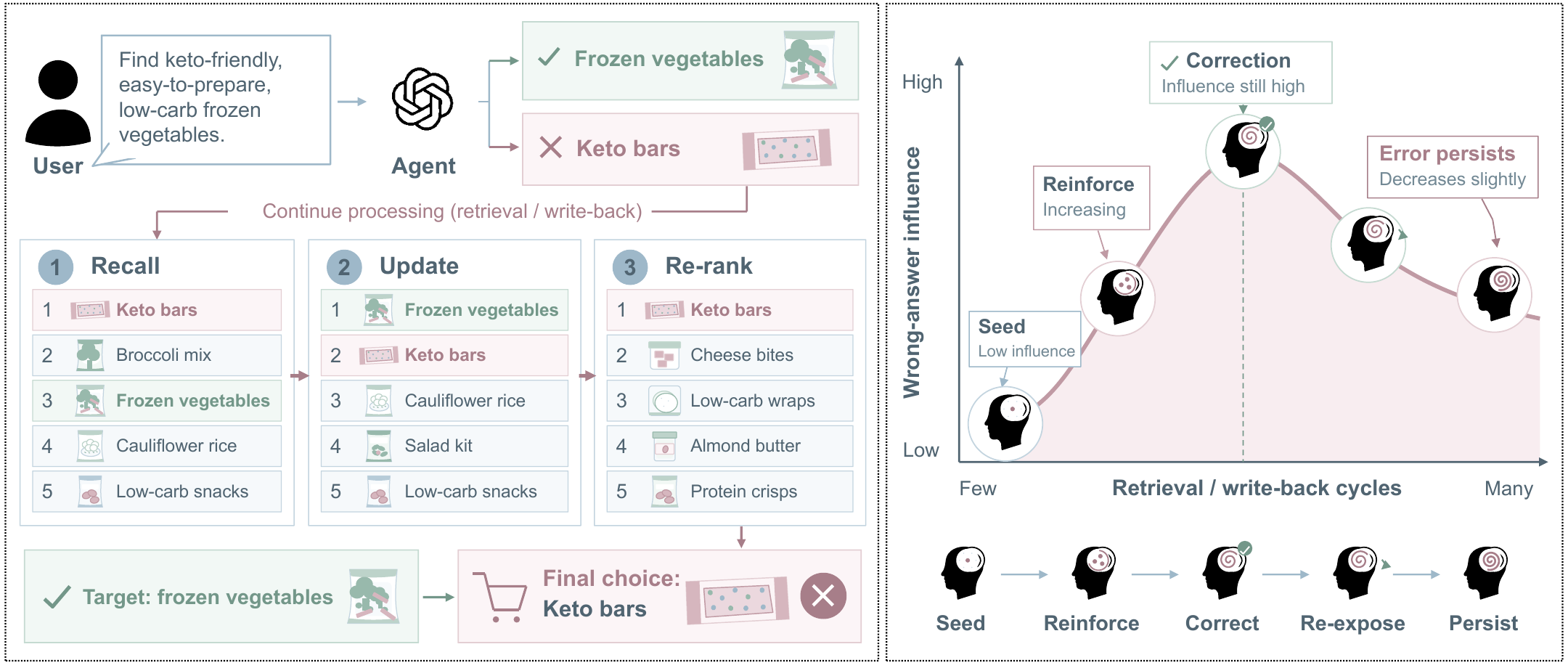}
\caption{Memory Echo in a shopping query. Repeated write-back gives one historical product several retrieval positions, so its descendants compete with the current correction and can redirect the final choice.}
\label{fig:echo-mechanism}
\vspace{-15pt}
\end{figure}

Long-term memory lets agents carry information across sessions, but current evidence should still determine present answers. A once-suitable product recommendation can become inappropriate, a medical answer can be corrected, and a travel itinerary can become obsolete. The issue is not only whether an old record remains stored: repeated retrieval, summarization, and write-back can create multiple retrievable traces from one source. Historical information can then occupy several positions in later queries and compete with a correction without any new independent evidence.

Research on agent long-term memory has advanced along several fronts. A-MEM and Mem0 develop mechanisms for memory organization, retrieval, and updates \citep{ref2,ref3}; LoCoMo, LongMemEval, and BEAM evaluate memory question answering over long horizons \citep{ref4,ref5,ref7}. Multi-agent communication and its security risks have been surveyed \citep{kong2025agentcommunication}; MANBENCH studies collective false memories \citep{ref1}, while corrective retrieval addresses evidence quality \citep{ref12}. These lines of work leave open a lifecycle evaluation of how same-source memory accumulates through repeated retrieval and write-back within a single agent, remains exposed after correction, and ultimately affects current answers.

Research on socially transmitted memory \citep{ref1} and on repeated retrieval, reconsolidation, and retention \citep{ref8,ref9,ref10,ref11} motivates our examination of this process. We call the extra retrieval exposure that historical information from one source gains through repeated processing without additional independent evidence \emph{Memory Echo}. In a shopping pilot, repeated read-only retrieval leaves the recorded task scores unchanged, whereas writing interactions back increases historical occupancy and reduces answer quality. Final-answer accuracy alone cannot distinguish whether an error arises from trace proliferation, a correction failing to displace old traces, or a generator failing to use a retrieved correction.

We organize the study around four questions. \textbf{RQ1 (formation):} Across four memory architectures, how do seed-only storage, repeated read-only retrieval, and retrieval with write-back differ in same-source derivative count, historical Echo Density (ED), and old-answer adoption? \textbf{RQ2 (post-correction influence):} After a common verified correction, what remains in native retrieval versus served context, how does exposure persist through 30 read-only queries, and does it still lower answer accuracy? \textbf{RQ3 (FPA mechanism):} With matched candidates and a fixed generator, how much do Family, Provenance, and Authority each contribute beyond Baseline, Budget, CRAG-M, and PairLR, including the change from unconstrained to constrained answers? \textbf{RQ4 (transfer and robustness):} Do paired gains persist across four architectures, two generators, and shopping, medical, and travel tasks under provenance errors and incomplete corrections, while retaining valid history?

MEchoBench pairs source questions across memory histories and follows initialization, repetition, correction, delayed retrieval, and task answering. We track derivative counts and ED for RQ1; native and served retrieval, storage- and serving-side Reconsolidation Resistance Index (RRI), a 30-query ED curve, Windowed Echo Persistence (WEP), and final accuracy for RQ2; Decision Contamination Effect (DCE) and before/after answer changes for RQ3; and paired performance and valid-history retention for RQ4.

The formation and correction mechanisms in RQ1 and RQ2 motivate the Family--Provenance--Authority framework (FPA), and RQ3 asks which of its components account for the measured gains. FPA selects one retrieval representative per source family to limit the occupancy advantage of repeated traces, uses verified corrections to determine when old records may enter the current answer context, and applies authoritative evidence and answer constraints to reduce historical pull on the final output. Old records remain in storage, allowing storage and serving effects to be observed separately; RQ4 asks whether this balance survives changes in architecture, generator, task, and provenance quality.

We evaluate the memory lifecycle across four general data sources and compare FPA with three mechanisms on matched candidate pools. On 190 fixed source questions across four architectures, FPA raises DeepSeek accuracy from 0.825 to 0.920 and Kimi accuracy from 0.782 to 0.916; mean DCE falls from 0.106 to 0.030 and from 0.109 to 0.038, respectively. Shopping, medical question answering, and travel planning show how retrieval changes affect current answers. Our contributions are: \begin{itemize}
    \item We introduce Memory Echo as a source-level phenomenon and characterize how repeated processing gives historical information a retrieval advantage that can affect decisions.
    \item We develop MEchoBench and four measurements to track trace accumulation, correction resistance, delayed persistence, and answer shift within one memory lifecycle.
    \item We introduce FPA, combining family representative selection, post-correction eligibility, and authority-guided answering to reduce decision contamination while preserving traceable history, and validate its effects across general data sources and practical tasks.
\end{itemize}

\section{Related Work}\label{sec:related}
\subsection{Long-Term Memory Evaluation}
External memory gives an agent access to information beyond its prompt. A-MEM organizes linked notes, while Mem0 supports storing, searching, and updating memories \citep{ref2,ref3}. LoCoMo, LongMemEval, and BEAM test long-history question answering \citep{ref4,ref5,ref7}; HotpotQA tests evidence combination \citep{ref6}. MemoryAgentBench additionally evaluates incremental learning and selective forgetting \citep{hu2025memoryagentbench}. \citet{li2026relevancesufficiencyactuallycloses} study whether retrieved memories jointly contain sufficient evidence to answer a question. Echo asks whether repeated processing makes one source appear to be several independent pieces of evidence. Among industry-backed agent-memory systems, LangChain's LangMem is a prominent open-source project that provides tools for memory creation, search, and background consolidation~\citep{ref21}.

Work on retrieval practice and memory retention motivates our sequence of access, revision, and delayed observation \citep{ref8,ref9,ref10,ref11}. Here these are software events: searches, writes, and later retrievals. Corrective RAG assesses evidence quality \citep{ref12}, while learning-to-rank changes candidate order \citep{ref19}. Our CRAG-M comparator grades stored memories, and PairLR uses a linear pairwise logistic ranker. Neither identifies candidates descended from the same claim; Echo and FPA track that shared ancestry.

\section{Preliminaries}\label{sec:prelim}
\paragraph{Motivation}
In \texttt{shopping-0028}, the user requests low-carb frozen vegetables. Write-back multiplies an older keto-bar recommendation; its descendants crowd the correction and redirect the choice (Figure~\ref{fig:echo-mechanism}). Read-only search leaves storage unchanged, isolating the effect of write-back.

\paragraph{Agent Memory and Evidence Provenance}
At turn $t$, retriever $\mathcal R$ searches memory $M_t$ for query $q_t$; serving rule $\mathcal G$ chooses the answer context, and updater $\mathcal U$ changes storage. Each record carries content, time, source and family IDs, a parent pointer, validity, and exposure counts. A retrieval-derived paraphrase inherits its parent's source: several records can be one piece of evidence.

\paragraph{Operational Definition of Memory Echo}
Memory Echo is the extra retrieval exposure a historical claim and its descendants gain without independent evidence. The original claim need not have been false: the failure occurs when it keeps occupying evidence positions after the task changes or a verified correction arrives. We separate record from source counts and retrieval from answer bias.

\section{Agent Memory Echo}\label{sec:echo}\label{sec:benchmark}
One source becomes several records; correction fails to displace them; exposure persists into an answer. RQ1--RQ3 trace this lifecycle; RQ4 tests transfer across architectures, generators, and tasks.

\subsection{The MEchoBench Lifecycle}
\paragraph{Questions and historical seeds.}
We take 50 questions each from LoCoMo, LongMemEval, HotpotQA, and BEAM~\citep{ref4,ref5,ref6,ref7}, reserving ten LoCoMo questions for development and reporting 190. We retain wording, gold answer, supporting evidence, ability type, and source ID. A different real answer from the same dataset, preferably matching ability type, seeds history; donor IDs and derivative parent links preserve ancestry. The target gold answer supplies the verified correction.

\paragraph{Paired histories and five stages.}
For each question, three histories share the seed, neutral competitors, query sequence, correction, and retrieval budget. \emph{Once} stores the seed once; \emph{Read} retrieves it six times without writing; \emph{Write} follows each retrieval with a same-family derivative write. Encoding records the initial state, retrieval practice changes exposure, and updating inserts the same correction in all histories. Thirty subsequent read-only probes measure persistence without further writes; final retrieval supplies the answer context. Figure~\ref{fig:lifecycle-overview} maps these five stages to observations, with probe locations in Appendix Table~\ref{tab:stages}.

\paragraph{Lifecycle readouts.}
At encoding, we measure how much of the initial retrieval is occupied by the seed. After retrieval practice, we count its descendants and their share of the pre-correction retrieval. Once the correction is added, we check whether it is retrieved and how much of the old family remains in storage and in the returned list. We then run 30 read-only probes, recording retrieval scores without changing memory. Finally, we record the answer produced from the candidates served to the generator. These measurements distinguish an old claim that remains stored from one that is retrieved, and one that the answer actually adopts.

\begin{figure}[t]
\centering
\includegraphics[width=0.49\textwidth]{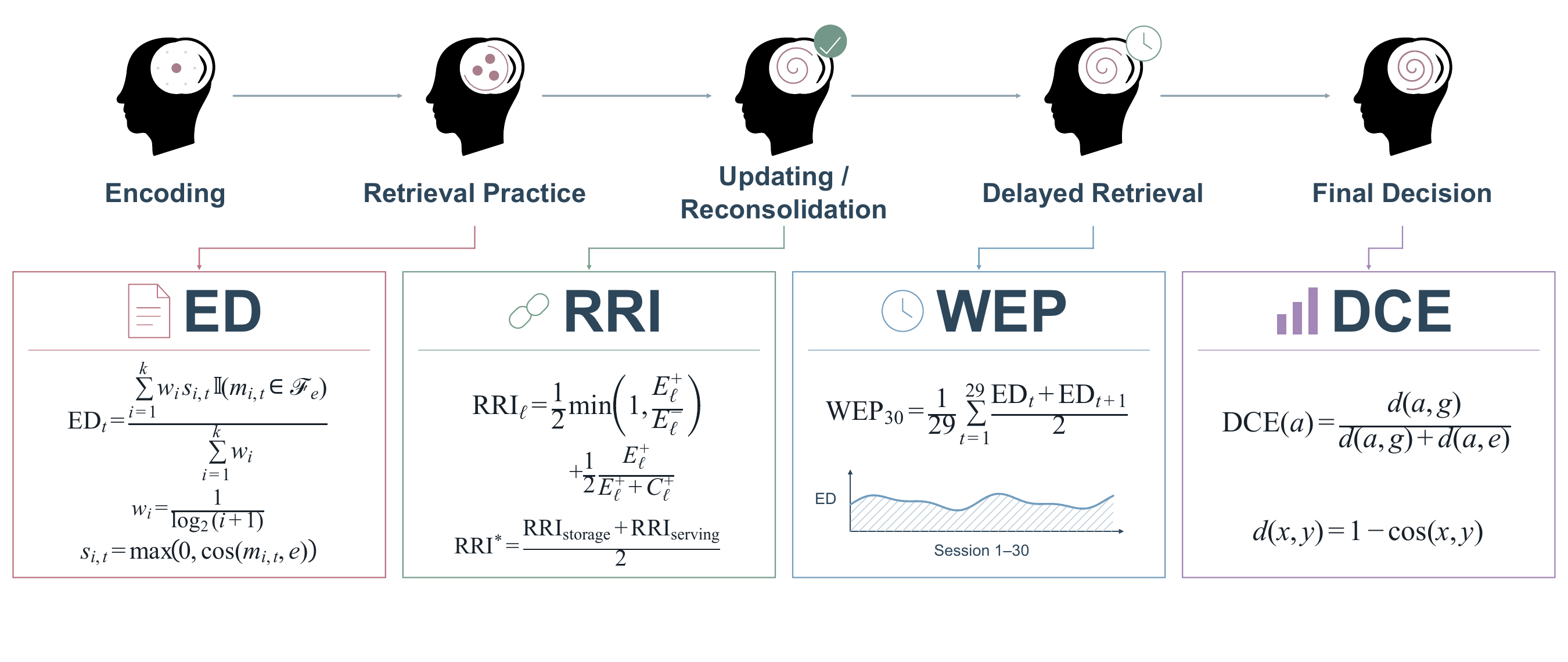}
\caption{MEchoBench's five stages. E-minus/E-plus mark old exposure before/after correction; C-plus marks corrected exposure. Here a/g/e denote answer/gold/Echo texts, and d is cosine distance. ED, RRI, WEP, and DCE use Equations~\ref{eq:ed}--\ref{eq:dce}; the pictured RRI and DCE formulas are schematic.}
\label{fig:lifecycle-overview}
\vspace{-15pt}
\end{figure}

\paragraph{What remains matched.}
We keep the final query and gold answer fixed across comparisons. The system first forms its native memory and retrieves candidates; we save both before replaying each serving method on the same candidate pool. This separates the records the system creates from the evidence the generator actually sees. Questions from the general datasets test source-level Echo, while the public-data shopping, medical-QA, and travel groups test whether it carries over to current tasks~\citep{ref15,ref16,ref17}. The experiments cover four memory architectures and two answer generators (Section~\ref{sec:experiments}). Because the memory trajectory stays fixed during replay, a change in the served context is not mistaken for a change in native memory.

\subsection{Metric Definitions}
\paragraph{Echo Density (ED).}
ED measures how much of the ranked answer context is occupied by one historical source family:
\begin{equation}\label{eq:ed}
\ED=\frac{\sum_{r=1}^{N}w_rb_rs_r}{\sum_{r=1}^{N}w_r}.
\end{equation}
\noindent\textit{Symbols.} $N$ is the number of retrieved records; $r$ is a record's rank from $1$ (highest) to $N$. $b_r$ is $1$ if the record at rank $r$ descends from the historical seed and $0$ otherwise. $x_r$ and $e$ are semantic representations of that record and the seed; $s_r=\max(0,\cos(x_r,e))$ is their nonnegative cosine similarity. $w_r=1/\log_2(r+1)$ is the rank weight, with $\log_2$ denoting the base-two logarithm. The sums add contributions over all ranks. Thus the numerator counts similarity-weighted old-family positions, and the denominator normalizes by all rank weights. High ED means prominent exposure, not necessarily a wrong answer.

\paragraph{Reconsolidation Resistance Index (RRI).}
RRI asks whether a verified correction displaces the old family. It combines the family's survival with its competition against the corrected evidence:
\begin{equation}\label{eq:rri}
\begin{aligned}
\RRI={}&\tfrac12\min(1,\ED_{\rm post}/\ED_{\rm pre})\\
&+\tfrac12\frac{\ED_{\rm post}}{\ED_{\rm post}+C_{\rm post}}.
\end{aligned}
\end{equation}
\noindent\textit{Symbols.} $\ED_{\rm pre}$ and $\ED_{\rm post}$ are old-family occupancies before and after correction; $C_{\rm post}$ is corrected-family occupancy afterward. Retrieval RRI uses rank-weighted occupancies of returned records; storage RRI uses unranked occupancies of stored records. Each family's semantic similarity is measured against its own seed or correction. $\min(1,\cdot)$ caps survival at one; the second fraction is the old share of competing evidence. The two $1/2$ factors give both parts equal weight, yielding RRI in $[0,1]$. RRI is zero when $\ED_{\rm pre}=0$; a zero denominator in the second fraction also makes that fraction zero.

\paragraph{Windowed Echo Persistence (WEP).}
WEP summarizes old-family exposure across the consecutive read-only queries after correction:
\begin{equation}\label{eq:wep}
\WEP=\frac{1}{29}\sum_{t=1}^{29}\frac{\ED_t+\ED_{t+1}}{2}.
\end{equation}
\noindent\textit{Symbols.} $\ED_t$ is old-family ED at read-only probe $t$, and $\ED_{t+1}$ is its value at the next probe. There are 30 probes; the sum runs over $t=1,\ldots,29$ to cover their 29 adjacent intervals. The index counts queries, not elapsed time. Each adjacent pair is averaged by dividing by $2$, and division by $29$ averages across the full window. $\WEP$ therefore measures persistence even when ED fluctuates. No probe writes a record.

\paragraph{Decision Contamination Effect (DCE).}
DCE tests whether the generated answer aligns more with the historical claim than with the verified answer:
\begin{equation}\label{eq:dce}
\DCE=\max(0,(E-G)/4).
\end{equation}
\noindent\textit{Symbols.} $a$ is the generated answer. $E,G\in[0,4]$ are its judge-assigned alignment scores with the historical Echo and verified gold answers, respectively; zero means no alignment and four means full alignment. Dividing $E-G$ by $4$ normalizes their difference, and $\max(0,\cdot)$ clips negative values to zero. Thus $\DCE\in[0,1]$ measures the answer's extra alignment with the old claim.

\paragraph{Reading the metrics together.}
ED locates old-family occupancy, RRI its resistance to correction, WEP its delayed exposure, and DCE its influence on the answer. Reporting each at its stage separates a crowded retrieval list from a contaminated decision.
Correctness is judged against the verified answer, without showing the Echo, on a 0--4 scale. Scores of at least 3 count as accurate; a separate binary judgment records old-answer adoption.

\section{Provenance-Guided Mitigation}\label{sec:method}
\subsection{The Three-Stage FPA Pipeline}
\begin{figure*}[t]
\centering
\includegraphics[width=\textwidth]{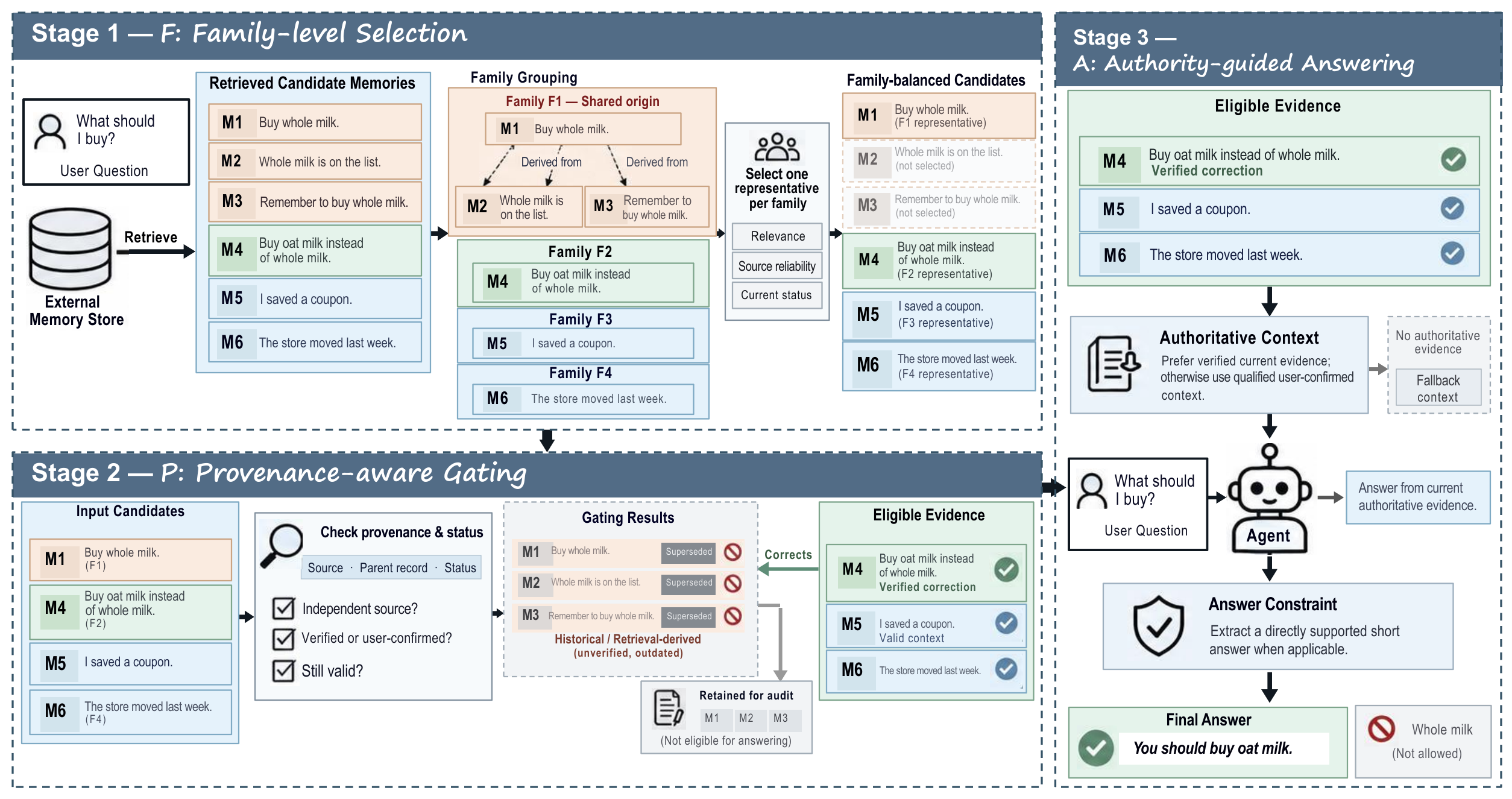}
\caption{Three-stage FPA pipeline: one serving position per source family, provenance-based eligibility, and an answer grounded in current authority. Superseded records remain auditable; the fallback box is illustrative, not an evaluated recovery rule.}
\label{fig:fpa-pipeline}
\vspace{-15pt}
\end{figure*}

\FPA{} controls the answer context between retrieval and generation without changing stored history. Records carry source and family IDs, parent links, source type, and validity; derivatives inherit their parent's family. Figure~\ref{fig:fpa-pipeline} shows family selection, provenance gating, and authority-guided answering in a shopping query.

\paragraph{Stage 1: Family-level selection.}
Source and parent links group the old whole-milk records M1--M3 in Figure~\ref{fig:fpa-pipeline} into family F1; M4--M6 have other origins. FPA scores retrieval relevance, confidence, source reliability, exposure, and validity (Appendix~\ref{app:repro}), then admits at most one representative per family. M1 represents F1 in the illustration, while M2 and M3 receive no extra positions. Repeated writes therefore add records, not independent votes.
The family ID follows ancestry, not the wording of the answer: two independent sources may agree, while two differently worded summaries of one retrieval remain the same family. The representative is selected anew from the current candidate pool; source grouping does not delete the unselected records.

\paragraph{Stage 2: Provenance-aware gating.}
The gate checks source, parent, verification, and validity against the correction. In the figure, M4 supersedes the whole-milk family: M1--M3 remain stored for audit but cannot support the current answer. M4 enters as verified evidence; nonconflicting M5 and M6 remain eligible context. Retrieval neither grants a record write-back authority nor restores an obsolete descendant's evidential status. Section~\ref{sec:rq4} tests this gate when correction quality changes.
This is the handoff from family balance to evidence validity. Selecting only M1 from F1 would still leave one obsolete whole-milk recommendation in the prompt; the provenance check removes that last route into current evidence after M4 arrives. Serving eligibility changes while the full record lineage remains available.

\paragraph{Stage 3: Authority-guided answering.}
The generator sees eligible evidence rather than raw Top-$k$ and is directed to prefer verified current facts. Here M4 supports ``You should buy oat milk,'' not the superseded whole-milk claim. A final check compares the draft with admitted authority and may select a supported short span for an extractive question; both raw and constrained answers are retained. The figure's no-authority fallback to qualified context is illustrative. The missing/noisy-correction experiment tests the gate without implementing that recovery branch.

\paragraph{Serving comparison.}
All methods serve at most five records from the same frozen pool. Budget penalizes excess old occupancy; CRAG-M grades candidates~\citep{ref12}; PairLR learns pairwise ranking~\citep{ref19}. FPA additionally controls family positions, post-correction eligibility, and the final answer's authority. Appendix~\ref{app:repro} gives the protocols.

\section{Experiments}\label{sec:experiments}
\subsection{Experimental Setup}
The lifecycle runs in A-MEM~\citep{ref2}, official Mem0~\citep{ref3}, CoM~\citep{ref20}, and LangMem~\citep{ref21}. Two reinforced histories for each of 190 questions in four architectures yield 1,520 paired answer states per method and generator. DeepSeek V4 Flash uses temperature 0 and Kimi K2.6 uses 0.6; both produce at most 120 tokens. GPT-5.6 (\texttt{gpt-5.6-sol}) judges answers in two passes. Within each generator, methods share questions, histories, and candidate pools. ED and WEP depend on the served pool, so they match across generators. We compare methods within each model; Appendix~\ref{app:main-details} reports architecture cells.

\subsection{Metrics}
Accuracy uses the judge's correctness score, while old-answer rate records whether the generated answer adopts the historical claim. ED, RRI, WEP, and DCE retain the distinct roles defined in Section~\ref{sec:echo}. Methods are compared on paired question--architecture--history states. For uncertainty intervals, we first average states belonging to the same source question and then resample the 190 questions, so repeated states from one question do not become independent observations.

\subsection{Main Results}\label{sec:mainresults}
The lifecycle experiments locate where Echo enters the pipeline, and the main method comparison asks whether that influence can be interrupted when the source question, correction, and candidate pool remain fixed. Table~\ref{tab:main} reports both answer models; the stage-specific contrasts explain its results.

\paragraph{RQ1: Same-Source Retrieval Occupancy}\label{sec:rq1}
The first contrast separates remembering a claim from multiplying it. Repeated read-only search leaves pre-correction ED essentially unchanged from one-time encoding. When the same searches write derivatives back, old-family occupancy rises sharply in every architecture, and old answers reappear after the common correction (Appendix Table~\ref{tab:app-a}). The extra records all inherit one source: write-back amplifies access to a claim without adding independent support. Its cross-architecture consistency points to the retrieval--write feedback loop, not a particular ranker.

\paragraph{RQ2: Correction Resistance and Persistence}\label{sec:rq2}
All histories receive the same verified correction, yet write-back lets old descendants crowd it out of retrieval. Storage and retrieval RRI rise as descendants survive the update and compete for limited evidence slots. Appendix~\ref{app:lifecycle} shows this competition is strongest in CoM.

\begin{figure}[!t]
\centering
\includegraphics[width=\columnwidth]{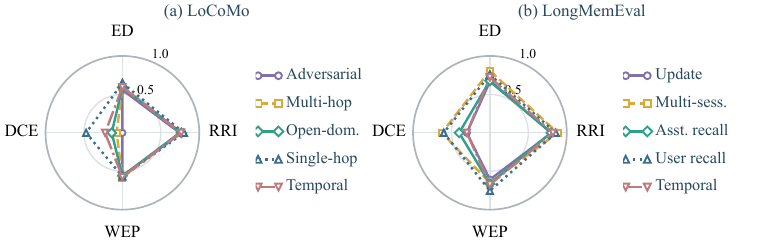}
\caption{Echo by ability in LoCoMo and LongMemEval. Exact values appear in Appendix Table~\ref{tab:source-abilities}.}
\label{fig:source-radars-current}
\vspace{-15pt}
\end{figure}

Stopping writes does not erase that advantage. The read-only ED values remain elevated and nearly flat across the delayed probes (Table~\ref{tab:app-c}); WEP therefore captures an inherited retrieval bias rather than ongoing reinforcement. The complete trajectories are in Appendix Tables~\ref{tab:app-b}--\ref{tab:app-c-30}. The verified answer is stored while descendants of the historical answer still fill the serving path (Appendix~\ref{app:refreshed-cases}). The ability-level radar shows that this exposure and its effect on answers vary by question type, so ED alone cannot stand in for decision contamination (Figure~\ref{fig:source-radars-current}; Appendix Table~\ref{tab:source-abilities}).
\begin{table*}[!t]
\centering\small\setlength{\tabcolsep}{3pt}
\begin{tabular*}{\textwidth}{@{\extracolsep{\fill}}lrrrrrrrr@{}}\toprule
Memory & Pre & Post & $t{=}1$ & $t{=}10$ & $t{=}20$ & $t{=}30$ & Old & Ret.\\\midrule
\mbox{A-MEM~\citep{ref2}} & 0.932 & 0.704 & 0.660 & 0.664 & 0.665 & 0.654 & 0.195 & 1.000\\
\mbox{Mem0~\citep{ref3}} & 0.932 & 0.704 & 0.660 & 0.664 & 0.665 & 0.654 & 0.211 & 1.000\\
\mbox{CoM~\citep{ref20}} & 0.932 & 0.722 & 0.681 & 0.687 & 0.684 & 0.671 & 0.232 & 1.000\\
\mbox{LangMem~\citep{ref21}} & 0.932 & 0.704 & 0.660 & 0.664 & 0.665 & 0.654 & 0.211 & 1.000\\\bottomrule
\end{tabular*}
\caption{DeepSeek persistence after write-back (190 questions per row). All 30 delay queries were read-only. Pre, Post, and the four query columns report ED; Old is the final old-answer rate and Ret. is the final old-retrieval proxy.}
\label{tab:app-c}
\end{table*}
\begin{figure}[!t]
\centering
\includegraphics[width=\columnwidth]{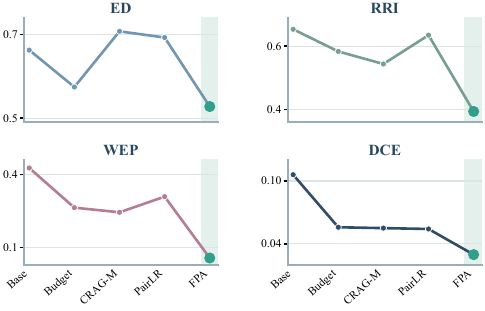}
\caption{DeepSeek method profiles for ED, RRI, WEP, and DCE; lower is better.}
\label{fig:method-metric-profile}
\vspace{-10pt}
\end{figure}

\paragraph{RQ3: Answers to Current Tasks}\label{sec:main}
The lifecycle establishes how an old family reaches the generator; Table~\ref{tab:main} asks whether serving can interrupt that path. \FPA{} improves accuracy and reduces old-answer adoption with both DeepSeek and Kimi. Their ED and WEP changes coincide because the intervention alters the shared evidence pool before either model generates an answer. DCE also falls in both blocks, connecting lower exposure to a weaker pull on the final response.

Budget, CRAG-M, and PairLR often improve accuracy without removing persistent old-family exposure. CRAG-M is especially revealing: a useful candidate can help an answer even while same-source descendants still occupy ranked positions. \FPA{} instead changes both how many positions a family receives and whether its obsolete members remain eligible after correction. The delayed-exposure contrast makes this mechanism clearer than a single search snapshot.

\subsection{Framework and Model Variation}
\begin{figure}[!ht]
\centering
\includegraphics[width=\columnwidth]{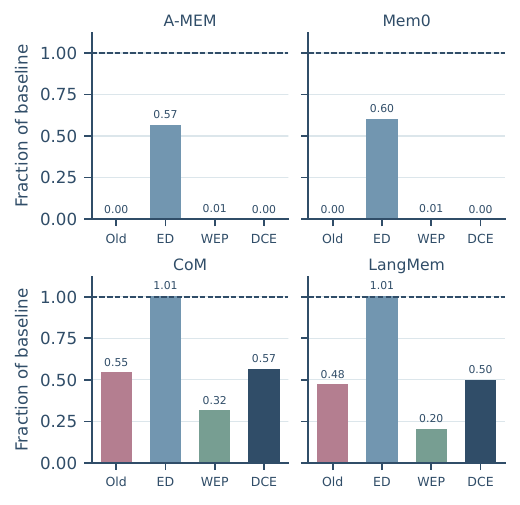}
\caption{DeepSeek V4 Flash risk under FPA relative to each architecture's Baseline (Table~\ref{tab:app-d-deepseek}). The dashed line marks Baseline$=1$; shorter bars indicate lower risk on the displayed metrics.}
\label{fig:architecture-risk}
\vspace{-13pt}
\end{figure}

The architecture breakdown reveals two regimes (Appendix~\ref{app:main-details}). A-MEM and Mem0 show large accuracy gains alongside near-elimination of delayed exposure. CoM and LangMem improve more modestly in accuracy, yet their WEP also falls: a crowded final retrieval does not imply sustained exposure across the delay window. Kimi follows the same broad split (Appendix Table~\ref{tab:app-d-kimi}). Question-level paired bootstrap intervals support positive accuracy changes and lower DCE for both generators. We resample source questions rather than their correlated architecture--history states, so the uncertainty reflects variation across questions rather than repeated views of the same one.
Figure~\ref{fig:architecture-risk} makes the architecture split visible relative to each Baseline. A-MEM and Mem0 nearly eliminate persistent exposure and decision contamination. CoM and LangMem show a different response: the final ED can remain close to Baseline while WEP and DCE fall. The intervention therefore changes repeated exposure and answer influence even where a single retrieval snapshot changes little.
\begin{table}[!t]
\centering\footnotesize\setlength{\tabcolsep}{3pt}
\begin{tabular*}{\columnwidth}{@{\extracolsep{\fill}}lrrrrr@{}}\toprule
Method & Acc.$\uparrow$ & Old$\downarrow$ & ED$\downarrow$ & WEP$\downarrow$ & DCE$\downarrow$\\\midrule
\multicolumn{6}{@{}l}{\emph{DeepSeek V4 Flash}}\\
Baseline & 0.825 & 0.106 & 0.662 & 0.427 & 0.106\\
Budget & 0.861 & 0.059 & 0.574 & 0.264 & 0.056\\
CRAG-M & 0.855 & 0.059 & 0.707 & 0.245 & 0.055\\
PairLR & 0.861 & 0.058 & 0.692 & 0.309 & 0.054\\
\FPA & \textbf{0.920} & \textbf{0.028} & \textbf{0.528} & \textbf{0.057} & \textbf{0.030}\\\midrule
\multicolumn{6}{@{}l}{\emph{Kimi K2.6}}\\
Baseline & 0.782 & 0.108 & 0.662 & 0.427 & 0.109\\
Budget & 0.847 & 0.075 & 0.574 & 0.264 & 0.073\\
CRAG-M & 0.842 & 0.064 & 0.707 & 0.245 & 0.064\\
PairLR & 0.845 & 0.068 & 0.692 & 0.309 & 0.065\\
\FPA & \textbf{0.916} & \textbf{0.038} & \textbf{0.528} & \textbf{0.057} & \textbf{0.038}\\\bottomrule
\end{tabular*}
\caption{\textbf{Main comparison.} Four-architecture means over 1,520 paired states per generator. Accuracy is higher-is-better; other metrics are lower-is-better. ED and WEP depend on the served pool.}
\label{tab:main}
\par\raggedright\scriptsize CRAG-M adapts \citet{ref12}; PairLR follows \citet{ref19}. Frameworks: A-MEM~\citep{ref2}, Mem0~\citep{ref3}, CoM~\citep{ref20}, LangMem~\citep{ref21}.
\vspace{-23pt}
\end{table}

Appendix Figure~\ref{fig:method-bars} compares shared context changes with the two generators' answer responses.

\subsection{Paired Mechanism Comparisons}
The matched comparisons show that \FPA{} couples answer improvement with a larger drop in persistent exposure than the three alternative serving rules. Kimi's paired accuracy gains are positive across all four frameworks (Appendix Table~\ref{tab:kimi-gains-main}), while its DCE ordering is unchanged. The paired intervals in Appendix Figure~\ref{fig:paired-current} separate this family-level exposure effect from a generic improvement in answer selection.
CRAG-M and PairLR can elevate useful candidates without excluding same-source descendants. \FPA{} changes eligibility before ranking, so its WEP advantage persists across the read-only delay. The paired intervals resample source questions, preserving the shared comparison across architectures and histories.

\subsection{Variation across Data Sources}\label{sec:rq4}
Task transfer depends on whether history actually conflicts with current evidence. In shopping and medical QA, \FPA{} reduces old-answer adoption while improving accuracy. Travel begins without old-answer adoption; restricting useful planning context then reduces accuracy (Table~\ref{tab:app-g}). The result identifies a boundary of the intervention, not a universal benefit from filtering history.

An earlier extractive pilot measured Post ED, RRI, WEP, and DCE on shopping, medical QA, and travel (Figure~\ref{fig:scenarios-legacy}). It used 50 tasks per scenario across A-MEM and Mem0. Its values are in Appendix Table~\ref{tab:legacy-scenarios}; this pilot is distinct from the current transfer study (Table~\ref{tab:app-g}).

\begin{figure}[!t]
\centering
\includegraphics[width=\columnwidth]{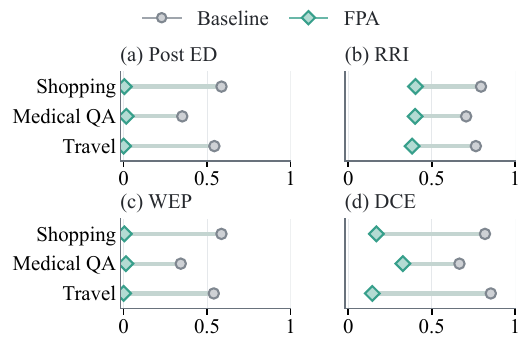}
\caption{Legacy three-scenario pilot. Rows show shopping, medical QA, and travel; circles are Baseline and diamonds are FPA. Exact values appear in Appendix Table~\ref{tab:legacy-scenarios}. The pilot used an earlier extractive protocol, not the current task-transfer cohort.}
\label{fig:scenarios-legacy}
\vspace{-10pt}
\end{figure}
Source-quality tests sharpen that boundary. Complete provenance and noisy family IDs retain a modest advantage, whereas a missing or unreliable correction leaves the gate without a trustworthy replacement and harms accuracy. When the old fact is nonconflicting, the change is small (Appendix Table~\ref{tab:app-h}). The contrast separates evidence validity from recency: filtering helps only when reliable current evidence replaces the old source.

\subsection{Ablation Results: Sequential Additions}\label{sec:ablation}
The ablation separates suppressing obsolete evidence from supplying a usable answer. Provenance control alone sharply reduces old-answer adoption but does not improve correctness; family control by itself can also remove useful context. Combining the two gates further reduces the historical claim's reach but leaves the generator short of a replacement answer. Accuracy recovers when verified authority is added, and the answer check contributes a smaller final gain (Table~\ref{tab:main-ablation}). Together, these stages limit unsupported repetition and supply the current fact.
\begin{table}[t]
\centering\small\setlength{\tabcolsep}{3pt}
\begin{tabular*}{\columnwidth}{@{\extracolsep{\fill}}lrrr@{}}\toprule
Variant & Acc. & Old & DCE\\\midrule
Baseline & 0.825 & 0.106 & 0.106\\
\textit{w} Family & 0.651 & 0.095 & 0.097\\
\textit{w} Provenance & 0.820 & 0.010 & 0.009\\
\textit{w} Joint gate & 0.637 & 0.012 & 0.011\\
\textit{w} Authority & 0.912 & 0.031 & 0.032\\
Full FPA & \textbf{0.920} & \textbf{0.028} & \textbf{0.030}\\\bottomrule
\end{tabular*}
\caption{DeepSeek component ablation on 1,520 paired states. The full component-presence matrix is in Appendix Table~\ref{tab:app-e}.}
\label{tab:main-ablation}
\vspace{-13pt}
\end{table}

\subsection{Representative Success and Failure}\label{sec:cases}
Untreated DeepSeek repeats the historical answer, whereas \FPA{} returns the verified one. The trace links that reversal to which source family reaches the generator, not to a changed question (Section~\ref{sec:rq2}). Travel and unreliable-correction cases show the converse: when the old answer is not being adopted or replacement evidence is unavailable, withholding context can remove information the generator still needs. The serving decision must therefore depend on a real conflict and a usable authority; Table~\ref{tab:correction-quality-main} illustrates the correction-quality boundary.

\begin{table}[!ht]
\centering\small
\begin{tabular*}{\columnwidth}{@{\extracolsep{\fill}}lrr@{}}
\toprule
Source condition & Baseline & \FPA{}\\
\midrule
Complete & 0.604 & 0.629\\
Correction missing/noisy & 0.378 & 0.212\\
\bottomrule
\end{tabular*}
\caption{DeepSeek accuracy by correction quality; 760 answer states per method and condition. Full results appear in Appendix Table~\ref{tab:app-h}.}
\label{tab:correction-quality-main}
\end{table}
\section{Conclusion}
We identify Memory Echo as a failure mode in external agent memory, where repeated write-back turns a single historical source into seemingly independent evidence. Across four architectures, this effect persists after correction and 30 unreinforced queries, but does not arise from read-only access. MEchoBench traces its propagation from source ancestry through retrieval to answers.

Our findings distinguish memory-side exposure from generator-dependent outcomes and reveal limitations of final retrieval snapshots. At the serving boundary, \FPA{} reduces Echo exposure and improves accuracy across two generators without discarding history. These results highlight a central principle for reliable agent memory: preserving historical information should not grant repeated records independent evidential authority, particularly after verified correction.

\section*{Limitations}
The benchmark places a real answer from another question into a controlled historical position; it does not estimate how frequently such seeds arise in natural interaction. Serving-method comparisons replay recorded candidate pools rather than evolving a separate memory database for each comparator. \FPA{} uses source-family and validity annotations supplied by the benchmark, and its final answer check can select a concise verified span, making part of the measured gain an answer-processing effect. GPT-5.6 judges correctness and contamination without an independent human audit of borderline cases. The travel and missing-correction conditions show that the measured benefit depends on a relevant Echo conflict and usable replacement evidence.

\section*{Ethical Considerations}
Medical questions are used as evaluation cases, not clinical advice. Historical seeds intentionally conflict with verified answers and belong in controlled benchmark traces. Source questions, derived trajectories, and logs should be redistributed in accordance with each dataset's license and privacy conditions.

\bibliography{references}

\clearpage
\appendix
\makeatletter
\setlength{\@fptop}{0pt}
\setlength{\@fpsep}{8pt plus 2pt}
\setlength{\@fpbot}{0pt plus 1fil}
\setlength{\@dblfptop}{0pt}
\setlength{\@dblfpsep}{8pt plus 2pt}
\setlength{\@dblfpbot}{0pt plus 1fil}
\makeatother
\setlength{\textfloatsep}{8pt plus 2pt minus 2pt}
\setlength{\floatsep}{6pt plus 2pt minus 2pt}
\section*{Appendix Contents}
\noindent\hyperref[app:lifecycle]{A. Lifecycle (Tables A--C)}\dotfill\pageref*{app:lifecycle}\par
\noindent\hyperref[app:main-details]{B. Main methods (Table D; Kimi gains)}\dotfill\pageref*{app:main-details}\par
\noindent\hyperref[app:components]{C. Ablations and Kimi (E--F)}\dotfill\pageref*{app:components}\par
\noindent\hyperref[app:transfer]{D. Task and source (G--H)}\dotfill\pageref*{app:transfer}\par
\noindent\hyperref[app:repro]{E. Reproducibility}\dotfill\pageref*{app:repro}\par
\noindent\hyperref[app:interpretation]{F. Lifecycle interpretation}\dotfill\pageref*{app:interpretation}\par
\noindent\hyperref[app:comparators]{G. Comparator protocols}\dotfill\pageref*{app:comparators}\par
\noindent\hyperref[app:additional-plots]{H. Supplementary visual comparisons}\dotfill\pageref*{app:additional-plots}\par
\noindent\hyperref[app:refreshed-cases]{I. Five-stage cases (25)}\dotfill\pageref*{app:refreshed-cases}\par
\noindent\hspace*{1em}\hyperref[app:refreshed-stage-1]{I.1 Encoding}\quad\hyperref[app:refreshed-stage-2]{I.2 Practice}\quad\hyperref[app:refreshed-stage-3]{I.3 Update}\par
\noindent\hspace*{1em}\hyperref[app:refreshed-stage-4]{I.4 Delay}\quad\hyperref[app:refreshed-stage-5]{I.5 Decision}\par
\medskip
\noindent The entries above link to the all appendix sections, including the refreshed case catalogue. The Overleaf package contains only sources needed to compile the paper; machine-readable experiment records are maintained separately.

\section{Full lifecycle measurements}\label{app:lifecycle}
The full condition breakdown separates three moments in the same Echo trajectory. Table~\ref{tab:app-a} shows how each history changes occupancy before correction and old-answer adoption afterward. Table~\ref{tab:app-b} locates the correction in storage and retrieval. Table~\ref{tab:app-c} samples the subsequent read-only curve; WEP includes every one of its 30 probes.

The Once and Read histories provide a direct control for exposure without additional records: both have the same ED and no derived write. The Write history adds six same-source records before the common correction. Its post-correction old-answer rates range from 0.195 to 0.232, so the retrieval change has an observable answer consequence. The storage RRI in Table~\ref{tab:app-b} then shows that correcting the original record does not remove its descendants; retrieval RRI reports how many still reach the ranked output.

\begin{table}[!ht]
\centering\small
\begin{tabularx}{\columnwidth}{@{}lXl@{}}\toprule
Stage & Probe purpose & QA count\\\midrule
Encoding & Is the historical seed retrievable? & 1\\
Retrieval practice & Six accesses plus one common pre-correction probe & 7\\
Updating & Does verified evidence displace the seed? & 1\\
Delayed retrieval & Does old exposure remain after writes stop? & 30\\
Final decision & Does the current answer follow old or new evidence? & 1\\\bottomrule
\end{tabularx}
\caption{Stage-specific probes; delay queries are consecutive and read-only.}
\label{tab:stages}
\par\raggedright\footnotesize Source questions: LoCoMo~\citep{ref4}, LongMemEval~\citep{ref5}, HotpotQA~\citep{ref6}, BEAM~\citep{ref7}.
\end{table}

\begin{table}[!ht]
\centering\small\setlength{\tabcolsep}{2pt}
\begin{tabularx}{\columnwidth}{@{}lXrrrr@{}}\toprule
Memory & History & ED & Derivatives & Old rate & Ret.\\\midrule
\AMEM & Once & 0.390 & 0 & 0.000 & 0.505\\
 & Read & 0.390 & 0 & 0.000 & 0.505\\
 & Write & 0.932 & 6 & 0.195 & 1.000\\
\MEMZERO & Once & 0.402 & 0 & 0.000 & 0.505\\
 & Read & 0.402 & 0 & 0.000 & 0.505\\
 & Write & 0.932 & 6 & 0.211 & 1.000\\
\COMM & Once & 0.390 & 0 & 0.000 & 0.505\\
 & Read & 0.390 & 0 & 0.000 & 0.505\\
 & Write & 0.932 & 6 & 0.232 & 1.000\\
\LANGMEM & Once & 0.390 & 0 & 0.000 & 0.505\\
 & Read & 0.390 & 0 & 0.000 & 0.505\\
 & Write & 0.932 & 6 & 0.211 & 1.000\\\bottomrule
\end{tabularx}
\caption{DeepSeek formation results (190 questions per row). ED is measured before correction; Derivatives counts write-back records, Old is the post-correction old-answer rate, and Ret. is the final old-retrieval proxy.}
\label{tab:app-a}
\memorysources
\end{table}

\begin{table}[!ht]
\centering\small\setlength{\tabcolsep}{3pt}
\setlength{\tabcolsep}{2pt}
\begin{tabularx}{\columnwidth}{@{}Xrrrrr@{}}\toprule
Memory & Corr. & RRI$_s$ & RRI$_r$ & WEP$_{30}$ & Acc.\\\midrule
\AMEM\ O/R & 1.000 & 0.519 & 0.454 & 0.190 & 0.984\\
\AMEM\ Write & 0.797 & 0.872 & 0.763 & 0.658 & 0.684\\
\MEMZERO\ O/R & 1.000 & 0.519 & 0.451 & 0.195 & 0.984\\
\MEMZERO\ Write & 0.797 & 0.872 & 0.763 & 0.658 & 0.663\\
\COMM\ O/R & 1.000 & 0.519 & 0.454 & 0.190 & 0.984\\
\COMM\ Write & 0.688 & 0.872 & 0.785 & 0.677 & 0.642\\
\LANGMEM\ O/R & 1.000 & 0.519 & 0.454 & 0.190 & 0.984\\
\LANGMEM\ Write & 0.797 & 0.872 & 0.763 & 0.658 & 0.674\\\bottomrule
\end{tabularx}
\caption{DeepSeek correction results. O/R combines the identical Once and Read outcomes; $s$ and $r$ denote storage and retrieval RRI. Acc. is final-answer accuracy.}
\label{tab:app-b}
\memorysources
\end{table}

\begin{table}[!ht]
\centering\footnotesize\setlength{\tabcolsep}{2pt}
\begin{tabular*}{\columnwidth}{@{\extracolsep{\fill}}lrrrr@{}}\toprule
Query & \AMEM & \MEMZERO & \COMM & \LANGMEM\\\midrule
1 & 0.660 & 0.660 & 0.681 & 0.660 \\
2 & 0.652 & 0.652 & 0.670 & 0.652 \\
3 & 0.653 & 0.653 & 0.671 & 0.653 \\
4 & 0.653 & 0.653 & 0.671 & 0.653 \\
5 & 0.655 & 0.655 & 0.673 & 0.655 \\
6 & 0.655 & 0.655 & 0.676 & 0.655 \\
7 & 0.661 & 0.661 & 0.681 & 0.661 \\
8 & 0.658 & 0.658 & 0.677 & 0.658 \\
9 & 0.667 & 0.667 & 0.688 & 0.667 \\
10 & 0.664 & 0.664 & 0.687 & 0.664 \\
11 & 0.658 & 0.658 & 0.675 & 0.658 \\
12 & 0.661 & 0.661 & 0.682 & 0.661 \\
13 & 0.662 & 0.662 & 0.681 & 0.662 \\
14 & 0.655 & 0.655 & 0.673 & 0.655 \\
15 & 0.651 & 0.651 & 0.669 & 0.651 \\
16 & 0.663 & 0.663 & 0.685 & 0.663 \\
17 & 0.659 & 0.659 & 0.680 & 0.659 \\
18 & 0.654 & 0.654 & 0.675 & 0.654 \\
19 & 0.665 & 0.665 & 0.682 & 0.665 \\
20 & 0.665 & 0.665 & 0.684 & 0.665 \\
21 & 0.656 & 0.656 & 0.670 & 0.656 \\
22 & 0.666 & 0.666 & 0.682 & 0.666 \\
23 & 0.664 & 0.664 & 0.685 & 0.664 \\
24 & 0.659 & 0.659 & 0.679 & 0.659 \\
25 & 0.653 & 0.653 & 0.671 & 0.653 \\
26 & 0.655 & 0.655 & 0.672 & 0.655 \\
27 & 0.658 & 0.658 & 0.677 & 0.658 \\
28 & 0.654 & 0.654 & 0.670 & 0.654 \\
29 & 0.658 & 0.658 & 0.674 & 0.658 \\
30 & 0.654 & 0.654 & 0.671 & 0.654 \\
\bottomrule
\end{tabular*}
\caption{All 30 consecutive read-only ED probes after write-back. Each value averages the same 190 source questions in an architecture; these values underlie WEP$_{30}$.}
\label{tab:app-c-30}
\memorysources
\end{table}

\section{Main experiment by architecture}\label{app:main-details}
The pooled result conceals different starting points across memory systems. Table~\ref{tab:app-d-deepseek} places DeepSeek and Kimi side by side and retains all architecture--method cells, each averaging 380 answer states. Accuracy and old-answer rate come from generated text; ED and WEP describe the served memory. For Kimi, paired \FPA{} minus Baseline accuracy gains are 0.182 [0.145,0.218] on A-MEM, 0.179 [0.142,0.216] on Mem0, 0.097 [0.066,0.132] on CoM, and 0.082 [0.050,0.116] on LangMem (95\% source-level bootstrap intervals over 190 questions).

The architecture rows distinguish a change in final retrieval rank from a change in accumulated exposure. On CoM and LangMem, FPA's final ED is near their Baseline ED, yet WEP falls sharply. The complete 30-query curve therefore carries information that a single final search cannot provide. In A-MEM and Mem0, the answer check can use a concise verified span, producing larger accuracy gains. The side-by-side model blocks show that this split appears with both generators, while the magnitude of each gain remains architecture-dependent.
\begin{table}[!ht]
\centering
\footnotesize
\setlength{\tabcolsep}{2pt}
\renewcommand{\arraystretch}{1.05}
\begin{tabularx}{\columnwidth}{@{}lrr>{\raggedleft\arraybackslash}X@{}}
\toprule
Memory & Baseline & FPA & Gain [95\% CI]\\
\midrule
A-MEM   & 0.818 & 1.000 & 0.182 [0.145, 0.218]\\
Mem0    & 0.821 & 1.000 & 0.179 [0.142, 0.216]\\
CoM     & 0.729 & 0.826 & 0.097 [0.066, 0.132]\\
LangMem & 0.758 & 0.839 & 0.082 [0.050, 0.116]\\
\bottomrule
\end{tabularx}
\caption{Kimi K2.6 accuracy across memory architectures (190 source questions). Gain is FPA minus Baseline; brackets show 95\% bootstrap intervals. Architectures: A-MEM~\citep{ref2}, Mem0~\citep{ref3}, CoM~\citep{ref20}, and LangMem~\citep{ref21}.}
\label{tab:kimi-gains-main}
\end{table}
\begin{table*}[!t]
\centering\footnotesize\setlength{\tabcolsep}{2pt}
\begin{minipage}[t]{0.49\textwidth}
\centering\textbf{DeepSeek V4 Flash}\par\vspace{2pt}
\begin{tabular*}{\linewidth}{@{\extracolsep{\fill}}lrrrrr@{}}\toprule
Method & Acc. & Old & ED & WEP & DCE\\\midrule
\multicolumn{6}{@{}l}{\AMEM}\\
Baseline & 0.834 & 0.097 & 0.661 & 0.424 & 0.097\\
Budget & 0.945 & 0.018 & 0.474 & 0.111 & 0.013\\
CRAG-M & 0.932 & 0.021 & 0.749 & 0.109 & 0.016\\
PairLR & 0.934 & 0.021 & 0.749 & 0.233 & 0.016\\
FPA & 1.000 & 0.000 & 0.376 & 0.003 & 0.000\\\midrule
\multicolumn{6}{@{}l}{\MEMZERO}\\
Baseline & 0.824 & 0.105 & 0.667 & 0.427 & 0.105\\
Budget & 0.945 & 0.021 & 0.501 & 0.116 & 0.017\\
CRAG-M & 0.932 & 0.021 & 0.758 & 0.113 & 0.016\\
PairLR & 0.937 & 0.021 & 0.758 & 0.234 & 0.016\\
FPA & 1.000 & 0.000 & 0.403 & 0.003 & 0.000\\\midrule
\multicolumn{6}{@{}l}{\COMM}\\
Baseline & 0.813 & 0.116 & 0.661 & 0.434 & 0.116\\
Budget & 0.768 & 0.103 & 0.661 & 0.423 & 0.100\\
CRAG-M & 0.771 & 0.097 & 0.661 & 0.388 & 0.095\\
PairLR & 0.784 & 0.097 & 0.631 & 0.394 & 0.095\\
FPA & 0.832 & 0.063 & 0.666 & 0.138 & 0.066\\\midrule
\multicolumn{6}{@{}l}{\LANGMEM}\\
Baseline & 0.829 & 0.105 & 0.661 & 0.424 & 0.105\\
Budget & 0.787 & 0.095 & 0.661 & 0.407 & 0.093\\
CRAG-M & 0.787 & 0.095 & 0.661 & 0.372 & 0.093\\
PairLR & 0.787 & 0.092 & 0.631 & 0.377 & 0.089\\
FPA & 0.850 & 0.050 & 0.666 & 0.087 & 0.053\\\bottomrule
\end{tabular*}
\end{minipage}\hfill
\begin{minipage}[t]{0.49\textwidth}
\centering\textbf{Kimi K2.6}\par\vspace{2pt}
\begin{tabular*}{\linewidth}{@{\extracolsep{\fill}}lrrrrr@{}}\toprule
Method & Acc. & Old & ED & WEP & DCE\\\midrule
\multicolumn{6}{@{}l}{\AMEM}\\
Baseline & 0.818 & 0.100 & 0.661 & 0.424 & 0.100\\
Budget & 0.955 & 0.021 & 0.474 & 0.111 & 0.017\\
CRAG-M & 0.945 & 0.021 & 0.749 & 0.109 & 0.016\\
PairLR & 0.942 & 0.024 & 0.749 & 0.233 & 0.019\\
FPA & 1.000 & 0.000 & 0.376 & 0.003 & 0.000\\\midrule
\multicolumn{6}{@{}l}{\MEMZERO}\\
Baseline & 0.821 & 0.100 & 0.667 & 0.427 & 0.101\\
Budget & 0.947 & 0.021 & 0.501 & 0.116 & 0.018\\
CRAG-M & 0.939 & 0.029 & 0.758 & 0.113 & 0.026\\
PairLR & 0.942 & 0.026 & 0.758 & 0.234 & 0.022\\
FPA & 1.000 & 0.000 & 0.403 & 0.003 & 0.000\\\midrule
\multicolumn{6}{@{}l}{\COMM}\\
Baseline & 0.729 & 0.121 & 0.661 & 0.434 & 0.121\\
Budget & 0.734 & 0.134 & 0.661 & 0.423 & 0.139\\
CRAG-M & 0.737 & 0.108 & 0.661 & 0.388 & 0.112\\
PairLR & 0.742 & 0.111 & 0.631 & 0.394 & 0.105\\
FPA & 0.826 & 0.076 & 0.666 & 0.138 & 0.081\\\midrule
\multicolumn{6}{@{}l}{\LANGMEM}\\
Baseline & 0.758 & 0.111 & 0.661 & 0.424 & 0.115\\
Budget & 0.753 & 0.124 & 0.661 & 0.407 & 0.118\\
CRAG-M & 0.747 & 0.100 & 0.661 & 0.372 & 0.104\\
PairLR & 0.755 & 0.113 & 0.631 & 0.377 & 0.114\\
FPA & 0.839 & 0.074 & 0.666 & 0.087 & 0.073\\\bottomrule
\end{tabular*}
\end{minipage}
\caption{Main-method results by architecture: DeepSeek V4 Flash (left) and Kimi K2.6 (right). Each architecture--method cell contains 380 reinforced-history states.}
\label{tab:app-d-deepseek}\label{tab:app-d-kimi}
\par\vspace{2pt}\raggedright\footnotesize \textsuperscript{1} A-MEM~\citep{ref2}; \textsuperscript{2} Mem0~\citep{ref3}; \textsuperscript{3} CoM~\citep{ref20}; \textsuperscript{4} LangMem~\citep{ref21}; CRAG-M~\citep{ref12}; PairLR~\citep{ref19}. Retrieved-memory measurements are identical across the two generators.
\end{table*}

\begin{figure*}[!t]
\centering
\includegraphics[width=\textwidth]{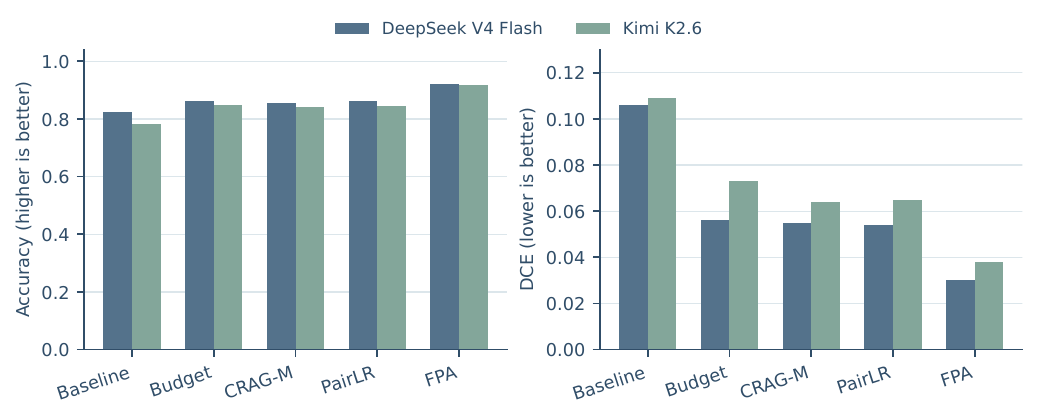}
\caption{Table D visualized by generator. Vertical bars compare accuracy and DCE for five serving methods on 1,520 paired answer states per model. DeepSeek and Kimi are shown separately; exact values appear in Table~\ref{tab:main}.}
\label{fig:method-bars}
\end{figure*}

\section{Components and cross-model gains}\label{app:components}
Table~\ref{tab:app-e} retains every DeepSeek component configuration, while Table~\ref{tab:kimi-gains-main} reports the paired Kimi Baseline-to-FPA change by architecture. Both use the same source questions as the main method table; the Kimi intervals resample 190 question-level differences.

The first two component rows isolate different tasks. Family control limits multiplicity, and provenance removes old-answer eligibility, but neither by itself gives the generator the replacement fact. Adding authoritative evidence to Family with Provenance changes accuracy from 0.637 to 0.912; the final answer check adds a smaller further gain. Table~\ref{tab:kimi-gains-main} then tests whether the complete pipeline's accuracy effect survives a change of generator rather than assigning model transfer to any single component.

\begin{table}[!ht]
\centering\small\setlength{\tabcolsep}{3pt}
\begin{tabularx}{\columnwidth}{@{}Xrrr@{}}\toprule
Configuration & Acc. & Old & DCE\\\midrule
Baseline & 0.825 & 0.106 & 0.106\\
Family only & 0.651 & 0.095 & 0.097\\
Provenance only & 0.820 & 0.010 & 0.009\\
Family w/ Provenance & 0.637 & 0.012 & 0.011\\
\quad w/ Authority & 0.912 & 0.031 & 0.032\\
\quad w/ Answer check (FPA) & 0.920 & 0.028 & 0.030\\\bottomrule
\end{tabularx}
\caption{Experiment E: DeepSeek FPA components; 1,520 paired states per configuration. Family and Provenance are tested separately before the components are added sequentially.}
\label{tab:app-e}
\par\raggedright\scriptsize This is an ablation of the FPA method; the compared retrieval and ranker families in Table~\ref{tab:main} follow \citet{ref12,ref19}.
\end{table}

\section{DeepSeek task and source tests}\label{app:transfer}
Task transfer and source quality test distinct conditions for the gate. Table~\ref{tab:app-g} asks whether historical-answer adoption is present in a new task; Table~\ref{tab:app-h} asks whether a usable correction accompanies the filtering rule. Every source-condition row uses the same 760 paired states. ``Valid history retained'' measures storage retention and equals 1.0 throughout these \FPA{} runs. Stored history can remain intact even when a record is not served to the generator.

The shopping and medical tasks begin with substantial old-answer adoption, and both gain accuracy as that rate falls. Travel begins with no old-answer adoption, so reducing access to historical planning context does not address a measured Echo; its accuracy decreases. The source test identifies a second boundary. Family-ID noise leaves some provenance uncertain but still permits a reliable correction to be used. When the correction itself is missing or noisy, the gate can withhold old context without supplying a valid replacement. The retained-history column confirms that this is a serving failure, not deletion of stored records.

\begin{table}[!ht]
\centering\footnotesize\setlength{\tabcolsep}{1.5pt}
\begin{tabular*}{\columnwidth}{@{\extracolsep{\fill}}lrrrrr@{}}\toprule
Task & BAcc & FAcc & BOld & FOld & $\Delta$Old\\\midrule
Shopping\textsuperscript{1} & 0.610 & 0.870 & 0.305 & 0.020 & -0.285\\
Medical QA\textsuperscript{2} & 0.585 & 0.850 & 0.395 & 0.040 & -0.355\\
Travel\textsuperscript{3} & 0.785 & 0.690 & 0.000 & 0.000 & 0.000\\\bottomrule
\end{tabular*}
\caption{DeepSeek task transfer (200 answer states per method and task). B and F denote Baseline and FPA; Acc is accuracy, Old is old-answer rate, and $\Delta$Old is FPA minus Baseline.}
\label{tab:app-g}
\par\raggedright\footnotesize \textsuperscript{1} WebShop~\citep{ref15}; \textsuperscript{2} MedQA~\citep{ref16}; \textsuperscript{3} TravelPlanner~\citep{ref17}.
\end{table}

\begin{table}[!ht]
\centering\small\setlength{\tabcolsep}{3pt}
\begin{tabularx}{\columnwidth}{@{}Xrrr@{}}\toprule
Source condition & Base acc. & FPA acc. & Retained\\\midrule
Complete & 0.604 & 0.629 & 1.000\\
Family ID noise & 0.599 & 0.630 & 1.000\\
Correction missing/noisy & 0.378 & 0.212 & 1.000\\
Old fact nonconflicting & 0.664 & 0.675 & 1.000\\\bottomrule
\end{tabularx}
\caption{DeepSeek source-quality test; 760 answer states per method and condition. FPA retains valid historical records in storage.}
\label{tab:app-h}
\par\raggedright\scriptsize Source questions: LoCoMo~\citep{ref4}, LongMemEval~\citep{ref5}, HotpotQA~\citep{ref6}, BEAM~\citep{ref7}.
\end{table}

\section{Reproducibility details}\label{app:repro}
The A--H CSVs and judged answers remain in the experiment's \texttt{results\_memory\_echo\_ah\_complete} and associated result directories, separate from this minimal Overleaf package. Accuracy uses correctness scores of at least 0.75; old-answer rate uses a separate binary contamination judgment, and DCE takes the positive Echo-minus-gold alignment difference. All 30 delay probes are read-only. Main-method means cover two reinforced histories per question and architecture, balanced across datasets and architectures after reserving ten LoCoMo development questions. The 25 selected case traces and all 30 aggregate delay means are reproduced in the appendix.

For a candidate memory record $m$, family representative selection uses
\begin{equation}\label{eq:fpa-score}
\begin{aligned}
S(m)={}&s_{\rm ret}(m)+0.35c(m)+0.80r(m)\\
&-0.18\log(1+v(m)+n(m))\\
&-4I_{\rm stale}(m),
\end{aligned}
\end{equation}

\noindent\textit{Symbols.} $m$ is a candidate memory record and $S(m)$ its score for selecting one representative from its source-ancestry family. $s_{\rm ret}(m)$ is the original retrieval score, $c(m)$ the record's confidence, and $r(m)$ its source reliability. $v(m)$ counts writes and $n(m)$ counts retrieval exposures. $\log$ is the natural logarithm, so the penalty for repeated writes and exposures grows sublinearly; the added $1$ makes the expression defined when both counts are zero. $I_{\rm stale}(m)$ equals $1$ for a superseded or suppressed record and $0$ otherwise. The numeric coefficients are fixed scoring weights, not probabilities. The highest-scoring eligible member takes the family's single serving position; its other records do not become independent evidence.

\section{Interpretation of the Echo lifecycle}\label{app:interpretation}
Memory Echo arises when one source claim becomes several retrievable records. These derivatives can crowd out a stored correction through 30 read-only probes. ED measures retrieval occupancy, RRI correction resistance, WEP delayed exposure, and DCE influence on the final answer. Occupancy need not cause an incorrect answer; suppressing history without replacement evidence can also fail.

\FPA{} caps family occupancy, gates superseded claims, and checks answers against authoritative evidence. Source-quality tests show the need for reliable replacement evidence; fixed-candidate replay isolates serving effects from native memory evolution.

\section{Comparator protocols and information access}\label{app:comparators}
Budget feeds occupancy above a Seed-only limit into a bounded feedback controller:
\begin{equation}\label{eq:budget-feedback}
h_t=\min(3,\,0.85h_{t-1}+1.25p_t).
\end{equation}

\noindent\textit{Symbols.} $f$ is the memory architecture, $s$ the lifecycle stage, and $b_{f,s}$ their Seed-only ED budget (the 95th percentile). At turn $t$, $E_t$ is old-family ED and $p_t=\max(0,E_t-b_{f,s})$ is its excess above budget. $h_{t-1}$ and $h_t$ are previous and current feedback strengths, initialized at zero. $0.85$ retains earlier pressure, $1.25$ weights new excess, and $\min(3,\cdot)$ caps the state at three. The state penalizes exposed or stale candidates and favors verified corrections; its signal requires labeled ED.

CRAG-M grades each retrieved candidate for reranking:
\begin{equation}\label{eq:cragm-quality}
Q(m)=0.45r+0.35v+0.15c-0.45u.
\end{equation}

\noindent\textit{Symbols.} $m$ is a retrieved candidate and $Q(m)$ its quality score, not a probability. $r\in[-1,1]$ is clipped retrieval relevance; $v$ is the verified-external-source indicator; $c\in[0,1]$ is confidence; $u$ is the conflict-or-stale indicator. Both indicators are binary. The fixed positive coefficients reward $r,v,c$, while the negative coefficient penalizes $u$. Low-quality candidates are demoted; absent reliable correction, a corrective-evidence flag is recorded. This adaptation of CRAG~\citep{ref12} omits its learned evaluator and web search.

PairLR learns a preference between a corrected candidate and an Echo candidate using retrieval, confidence, activation, write/exposure, provenance, and status features:
\begin{equation}\label{eq:pairlr}
P(m^+\succ m^-)=\sigma(\theta^\top\Delta\phi).
\end{equation}
\noindent\textit{Symbols.} $m^+$ is the corrected candidate and $m^-$ the Echo candidate; $\succ$ means ``should rank above.'' $\phi(m)$ is a candidate's feature vector, and $\Delta\phi=\phi(m^+)-\phi(m^-)$ is the corrected-minus-Echo feature difference. $\theta$ is the learned weight vector, $\theta^\top\Delta\phi$ their dot product, and $\sigma(z)=1/(1+e^{-z})$ the logistic function, with $z$ its input and $e$ the exponential base. $P$ is the predicted probability of preferring the corrected candidate. Five folds are split by source question, so each held-out question is scored without its own pair in training. PairLR reranks after correction without changing language-model weights; the comparison concerns candidate serving, not policy optimization.

\begin{table}[!htb]
\centering\footnotesize
\setlength{\tabcolsep}{2pt}
\renewcommand{\arraystretch}{0.94}
\begin{tabularx}{\columnwidth}{@{}l>{\raggedright\arraybackslash}X@{}}
\toprule
Method & Information used and serving operation\\
\midrule
\FPA{} & No Echo label for ranking; source-family IDs supplied. One position per family; verified correction supersedes history; authority guides and checks the answer.\\
Budget & Labeled ED sets the pressure and Seed-only budget. Exposure, conflict, and delay signals penalize excess occupancy.\\
CRAG-M & No Echo label in its quality score. Uses verified provenance, confidence, and stale status to rerank candidates and flag an absent correction.\\
PairLR & Labeled pairs for training only. Uses retrieval and provenance features for five-fold, question-held-out pairwise reranking.\\
\bottomrule
\end{tabularx}
\caption{Comparator information in paired candidate-pool replay. CRAG-M adapts \citet{ref12}; PairLR follows \citet{ref19}.}
\label{tab:app-access}
\end{table}

\section{Supplementary visual comparisons}\label{app:additional-plots}

\begin{figure}[!t]
\centering
\includegraphics[width=\columnwidth]{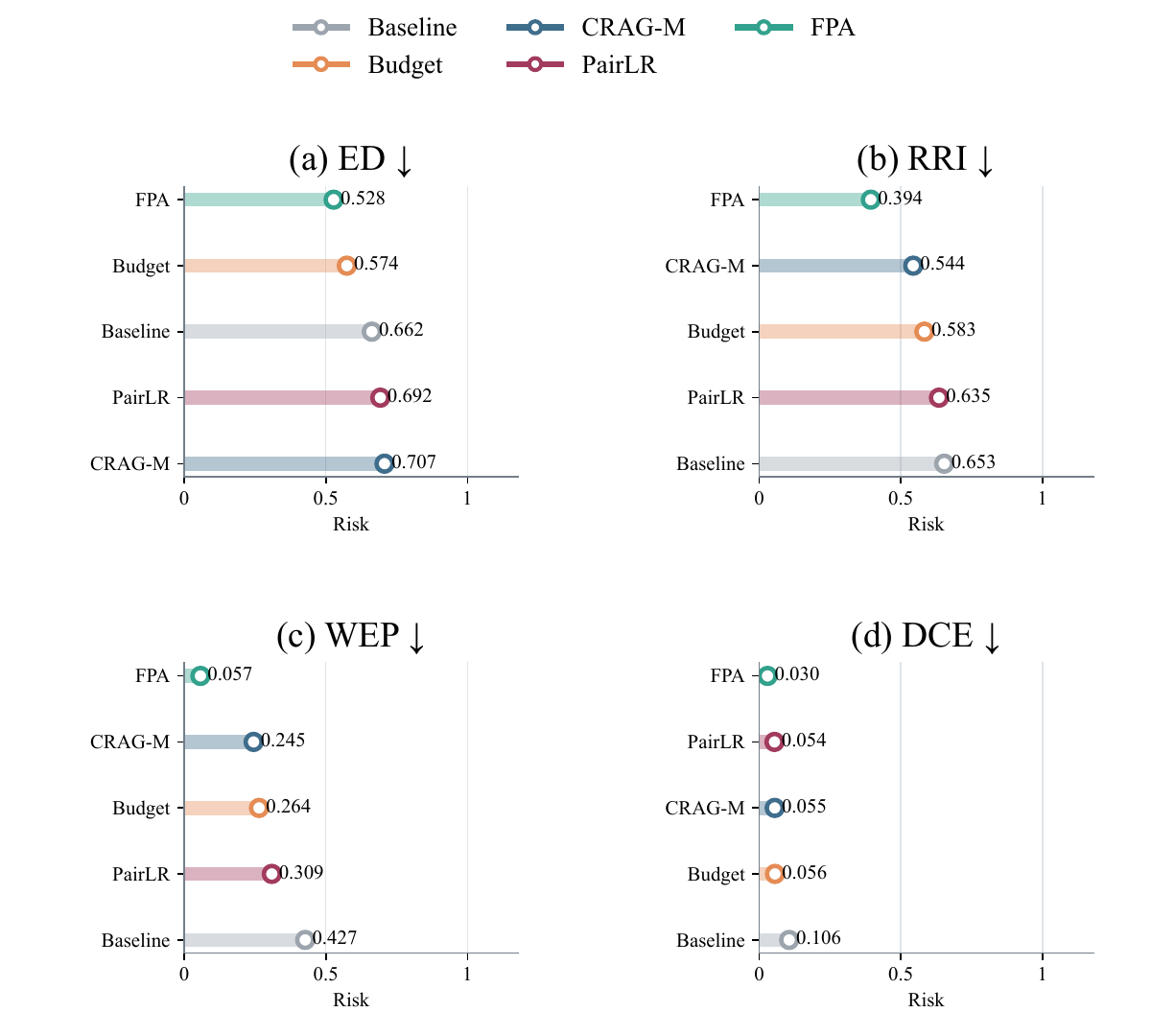}
\caption{Current DeepSeek method means for the four Echo measures, recomputed from the final 1,520-state Table D cohort. Lower is better in every panel; values are printed beside the marks.}
\label{fig:solutions-current}
\end{figure}

Figure~\ref{fig:solutions-current} shows the pooled DeepSeek levels behind the method contrasts. Figure~\ref{fig:paired-current} shows method-paired effects across the same question--architecture--history states as Table~\ref{tab:main}. ED, WEP, and DCE in the lollipop chart reproduce Table~\ref{tab:main}; RRI averages storage-side resistance from the matched lifecycle state and each method's serving-side resistance.

\noindent\begin{minipage}[t]{\columnwidth}
\centering
\setlength{\abovecaptionskip}{2pt}
\setlength{\belowcaptionskip}{1pt}

\includegraphics[width=0.90\columnwidth]{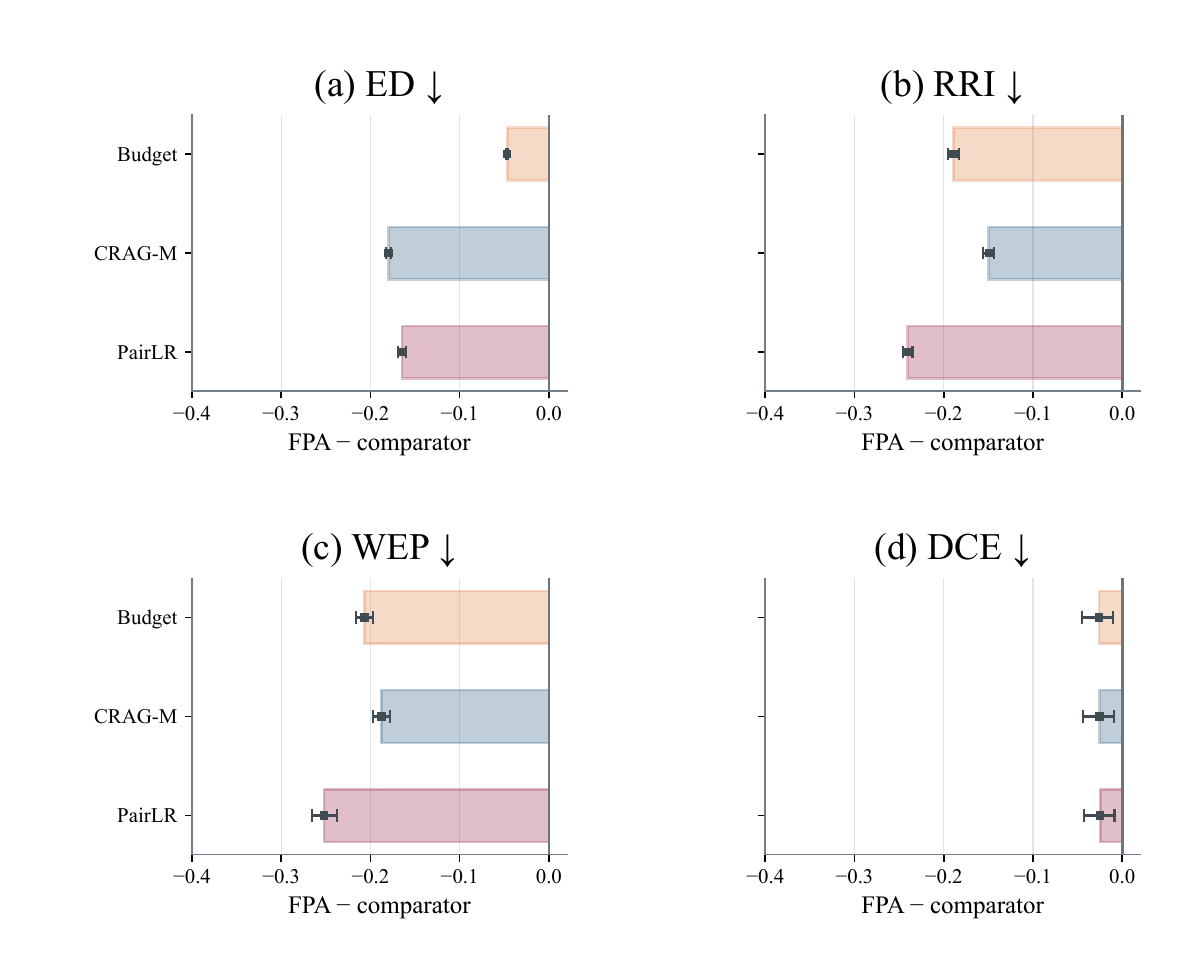}
\captionof{figure}{Paired DeepSeek effects on 190 questions, four architectures, and two write-back histories. Bars show mean FPA-minus-comparator effects; dark marks show 95\% source-question bootstrap intervals (10,000 resamples). Negative values favor FPA.}
\label{fig:paired-current}

\par\smallskip
{\footnotesize
\setlength{\tabcolsep}{1pt}
\renewcommand{\arraystretch}{0.90}
\begin{tabular*}{\columnwidth}{@{\extracolsep{\fill}}lrrrrr@{}}\toprule
Ability & Q & ED & RRI & WEP & DCE\\\midrule
\multicolumn{6}{@{}l}{\textit{LoCoMo}}\\
Adversarial & 2 & 0.558 & 0.745 & 0.563 & 0.000\\
Multi-hop & 10 & 0.600 & 0.760 & 0.581 & 0.075\\
Open-domain & 9 & 0.591 & 0.768 & 0.560 & 0.139\\
Single-hop & 9 & 0.646 & 0.800 & 0.568 & 0.472\\
Temporal & 10 & 0.593 & 0.765 & 0.575 & 0.225\\\midrule
\multicolumn{6}{@{}l}{\textit{LongMemEval}}\\
Update & 10 & 0.667 & 0.803 & 0.608 & 0.300\\
Multi-session & 10 & 0.806 & 0.882 & 0.652 & 0.625\\
Assistant recall & 10 & 0.678 & 0.795 & 0.662 & 0.400\\
User recall & 10 & 0.757 & 0.857 & 0.754 & 0.600\\
Temporal & 10 & 0.724 & 0.819 & 0.675 & 0.300\\\bottomrule
\end{tabular*}\par}
\captionof{table}{Ability-level means in Figure~\ref{fig:source-radars-current}. Q counts source questions; each contributes four framework states. Sources: LoCoMo~\citep{ref4} and LongMemEval~\citep{ref5}.}
\label{tab:source-abilities}
\end{minipage}

\begin{table}[!tb]
\centering\footnotesize
\setlength{\tabcolsep}{1pt}
\renewcommand{\arraystretch}{0.90}
\begin{tabular*}{\columnwidth}{@{\extracolsep{\fill}}llrrrr@{}}
\toprule
Legacy task & Method & Post ED & RRI & WEP & DCE\\
\midrule
Shopping & Baseline & 0.5864 & 0.7951 & 0.5861 & 0.8188\\
 & FPA & 0.0042 & 0.4025 & 0.0043 & 0.1683\\
Medical QA & Baseline & 0.3514 & 0.7060 & 0.3422 & 0.6658\\
 & FPA & 0.0157 & 0.4002 & 0.0137 & 0.3268\\
Travel & Baseline & 0.5436 & 0.7641 & 0.5401 & 0.8538\\
 & FPA & 0.0000 & 0.3820 & 0.0000 & 0.1436\\
\bottomrule
\end{tabular*}
\caption{Values for the earlier extractive pilot in Figure~\ref{fig:scenarios-legacy}, not the current Table G cohort. Tasks use WebShop~\citep{ref15}, MedQA~\citep{ref16}, and TravelPlanner~\citep{ref17}.}
\label{tab:legacy-scenarios}
\end{table}
\FloatBarrier
\section{Refreshed five-stage case catalogue}\label{app:refreshed-cases}
The previous appendix selected five illustrative cases for each lifecycle stage. We preserve that organization but recompute all retrieval counts, metrics, and recorded answers from the latest 190-question lifecycle and judged-answer files. The selected questions come from LoCoMo~\citep{ref4}, LongMemEval~\citep{ref5}, HotpotQA~\citep{ref6}, and BEAM~\citep{ref7}. The cards show the actual question and candidate answer strings, the retrieved records that made the historical claim visible, and the downstream generated answer. They are mechanism examples rather than prevalence estimates. Stage probes are retrieval-only; their associated Final Decision answer is a later observation from the same trajectory.

\noindent\emph{Data refresh.} The 25 case slots and five-stage grouping follow the previous appendix, but every question, retrieval, metric, and answer below is taken from the current 190-question \texttt{retrieve\_writeback} lifecycle and DeepSeek V4 Flash/GPT-5.6 judged-answer records. These are illustrative selections, not an incidence estimate. Each stage probe reads the same trajectory; only the Final Decision stage generates the recorded answer. RRI and WEP are trajectory-level measures, not additional per-probe judgments.
\noindent Four former LoCoMo source IDs fell within the ten development questions excluded from this run. Their case slots use same-dataset, same-ability held-out replacements: \texttt{multi-00025}$\to$\texttt{multi-00045}, \texttt{multi-00026}$\to$\texttt{multi-00046}, \texttt{multi-00027}$\to$\texttt{multi-00047}, and \texttt{multi-00029}$\to$\texttt{multi-00049}. No older case metric is mixed into the new results.\par

\subsection{Encoding}\label{app:refreshed-stage-1}

\subsubsection*{Case 1.1: multi-00068 $\mid$ LongMemEval $\mid$ Mem0}
\noindent\textbf{Ability and question.} single-session-assistant. I'm planning to revisit Orlando. I was wondering if you could remind me of that unique dessert shop with the giant milkshakes we talked about last time?\par
\noindent\textbf{Verified answer / historical seed.} The Sugar Factory at Icon Park. / I mentioned biometric authentication or one-time passwords (OTP) as examples of two-factor authentication methods.\par
\noindent\textbf{Stage probe.} Using stored memory, answer: I'm planning to revisit Orlando. I was wondering if you could remind me of that unique dessert shop with the giant milkshakes we talked about last time?\par
\noindent\textbf{Observed retrieval.} Top-4: 1 Echo-family and 0 correction record(s); stage ED 0.390.\par
\noindent\textbf{Top retrieved records.} \#1 ECHO: Historical memory answers the question 'I'm planning to revisit Orlando. I was wondering if you could remind me of that unique dessert shop with the giant milkshakes we talked about last time?' as: I mentioned biometric authentication or one-time passwords (OTP) as examples of two-factor authentication methods.; \#2 NOISE: Unrelated background record 2 for multi-00068; it contains no answer to the target question.; \#3 NOISE: Unrelated background record 1 for multi-00068; it contains no answer to the target question.\par
\noindent\textbf{Lifecycle measures.} Pre/post ED 0.976/0.541; retrieval RRI 0.584; WEP$_{30}$ 0.541; ED at delay 1/30 0.541/0.541.\par
\noindent\textbf{Downstream answer (same trajectory).} The Sugar Factory at Icon Park. Correctness 1.000; DCE 0.000; old-answer contamination: no.\par
\noindent\textbf{What this stage tests.} The seed's initial retrieval rank and ED show how much space it receives before practice.\par\medskip

\subsubsection*{Case 1.2: multi-00036 $\mid$ LoCoMo $\mid$ Mem0}
\noindent\textbf{Ability and question.} multi-hop. Would Caroline be considered religious?\par
\noindent\textbf{Verified answer / historical seed.} Somewhat, but not extremely religious / Pomodoro technique\par
\noindent\textbf{Stage probe.} Using stored memory, answer: Would Caroline be considered religious?\par
\noindent\textbf{Observed retrieval.} Top-4: 1 Echo-family and 0 correction record(s); stage ED 0.390.\par
\noindent\textbf{Top retrieved records.} \#1 ECHO: Historical memory answers the question 'Would Caroline be considered religious?' as: Pomodoro technique; \#2 NOISE: Unrelated background record 1 for multi-00036; it contains no answer to the target question.; \#3 NOISE: Unrelated background record 0 for multi-00036; it contains no answer to the target question.\par
\noindent\textbf{Lifecycle measures.} Pre/post ED 0.908/0.569; retrieval RRI 0.627; WEP$_{30}$ 0.569; ED at delay 1/30 0.569/0.569.\par
\noindent\textbf{Downstream answer (same trajectory).} Somewhat, but not extremely religious. Correctness 1.000; DCE 0.000; old-answer contamination: no.\par
\noindent\textbf{What this stage tests.} The seed's initial retrieval rank and ED show how much space it receives before practice.\par\medskip

\subsubsection*{Case 1.3: multi-00012 $\mid$ LoCoMo $\mid$ Mem0}
\noindent\textbf{Ability and question.} multi-hop. Would Caroline likely have Dr. Seuss books on her bookshelf?\par
\noindent\textbf{Verified answer / historical seed.} Yes, since she collects classic children's books / Middle-class or wealthy\par
\noindent\textbf{Stage probe.} Using stored memory, answer: Would Caroline likely have Dr. Seuss books on her bookshelf?\par
\noindent\textbf{Observed retrieval.} Top-4: 1 Echo-family and 0 correction record(s); stage ED 0.390.\par
\noindent\textbf{Top retrieved records.} \#1 ECHO: Historical memory answers the question 'Would Caroline likely have Dr. Seuss books on her bookshelf?' as: Middle-class or wealthy; \#2 NOISE: Unrelated background record 2 for multi-00012; it contains no answer to the target question.; \#3 NOISE: Unrelated background record 1 for multi-00012; it contains no answer to the target question.\par
\noindent\textbf{Lifecycle measures.} Pre/post ED 0.930/0.596; retrieval RRI 0.639; WEP$_{30}$ 0.596; ED at delay 1/30 0.596/0.596.\par
\noindent\textbf{Downstream answer (same trajectory).} Yes, since she collects classic children's books. Correctness 1.000; DCE 0.000; old-answer contamination: no.\par
\noindent\textbf{What this stage tests.} The seed's initial retrieval rank and ED show how much space it receives before practice.\par\medskip

\subsubsection*{Case 1.4: multi-00042 $\mid$ LoCoMo $\mid$ Mem0}
\noindent\textbf{Ability and question.} single-hop. What LGBTQ+ events has Caroline participated in?\par
\noindent\textbf{Verified answer / historical seed.} Pride parade, school speech, support group / A doll, a film camera\par
\noindent\textbf{Stage probe.} Using stored memory, answer: What LGBTQ+ events has Caroline participated in?\par
\noindent\textbf{Observed retrieval.} Top-4: 1 Echo-family and 0 correction record(s); stage ED 0.390.\par
\noindent\textbf{Top retrieved records.} \#1 ECHO: Historical memory answers the question 'What LGBTQ+ events has Caroline participated in?' as: A doll, a film camera; \#2 NOISE: Unrelated background record 2 for multi-00042; it contains no answer to the target question.; \#3 NOISE: Unrelated background record 1 for multi-00042; it contains no answer to the target question.\par
\noindent\textbf{Lifecycle measures.} Pre/post ED 0.919/0.564; retrieval RRI 0.619; WEP$_{30}$ 0.583; ED at delay 1/30 0.564/0.564.\par
\noindent\textbf{Downstream answer (same trajectory).} Pride parade, school speech, support group. Correctness 1.000; DCE 0.000; old-answer contamination: no.\par
\noindent\textbf{What this stage tests.} The seed's initial retrieval rank and ED show how much space it receives before practice.\par\medskip

\subsubsection*{Case 1.5: multi-00014 $\mid$ LoCoMo $\mid$ Mem0}
\noindent\textbf{Ability and question.} single-hop. Where did Caroline move from 4 years ago?\par
\noindent\textbf{Verified answer / historical seed.} Sweden / abstract art\par
\noindent\textbf{Stage probe.} Using stored memory, answer: Where did Caroline move from 4 years ago?\par
\noindent\textbf{Observed retrieval.} Top-4: 1 Echo-family and 0 correction record(s); stage ED 0.390.\par
\noindent\textbf{Top retrieved records.} \#1 ECHO: Historical memory answers the question 'Where did Caroline move from 4 years ago?' as: abstract art; \#2 NOISE: Unrelated background record 2 for multi-00014; it contains no answer to the target question.; \#3 NOISE: Unrelated background record 1 for multi-00014; it contains no answer to the target question.\par
\noindent\textbf{Lifecycle measures.} Pre/post ED 0.887/0.548; retrieval RRI 0.618; WEP$_{30}$ 0.587; ED at delay 1/30 0.548/0.548.\par
\noindent\textbf{Downstream answer (same trajectory).} Sweden Correctness 1.000; DCE 0.000; old-answer contamination: no.\par
\noindent\textbf{What this stage tests.} The seed's initial retrieval rank and ED show how much space it receives before practice.\par\medskip
\subsection{Retrieval Practice}\label{app:refreshed-stage-2}

\subsubsection*{Case 2.1: multi-00194 $\mid$ BEAM $\mid$ Mem0}
\noindent\textbf{Ability and question.} abstention. What was the atmosphere like during the February 20 book club discussion on 'The Poppy War' hosted by Kelly and I?\par
\noindent\textbf{Verified answer / historical seed.} Based on the provided chat, there is no information related to the atmosphere during the February 20 book club discussion on 'The Poppy War'. / Based on the provided chat, there is no information related to the topics or skills covered in the advanced storytelling workshop starting September 15.\par
\noindent\textbf{Stage probe.} Retrieve the historical memory and answer: What was the atmosphere like during the February 20 book club discussion on 'The Poppy War' hosted by Kelly and I?\par
\noindent\textbf{Observed retrieval.} Top-5: 5 Echo-family and 0 correction record(s); stage ED 0.982.\par
\noindent\textbf{Top retrieved records.} \#1 ECHO: Historical memory answers the question 'What was the atmosphere like during the February 20 book club discussion on 'The Poppy War' hosted by Kelly and I?' as: Based on the provided chat, there is no information related to the topics or skills covered in the advanced storytelling workshop starting September 15. Reconsolidation count 5.; \#2 ECHO: Historical memory answers the question 'What was the atmosphere like during the February 20 book club discussion on 'The Poppy War' hosted by Kelly and I?' as: Based on the provided chat, there is no information related to the topics or skills covered in the advanced storytelling workshop starting September 15. Retrieval-derived trace 5 repeats the historical answer for future use.; \#3 ECHO: Historical memory answers the question 'What was the atmosphere like during the February 20 book club discussion on 'The Poppy War' hosted by Kelly and I?' as: Based on the provided chat, there is no information related to the topics or skills covered in the advanced storytelling workshop starting September 15. Retrieval-derived trace 1 repeats the historical answer for future use.\par
\noindent\textbf{Lifecycle measures.} Pre/post ED 0.982/0.605; retrieval RRI 0.629; WEP$_{30}$ 0.605; ED at delay 1/30 0.605/0.605.\par
\noindent\textbf{Downstream answer (same trajectory).} There is no information available about the atmosphere during that book club discussion. Correctness 1.000; DCE 0.000; old-answer contamination: no.\par
\noindent\textbf{What this stage tests.} Six write-backs have occurred; compare this stage ED with pre-correction ED to locate amplification.\par\medskip

\subsubsection*{Case 2.2: multi-00186 $\mid$ BEAM $\mid$ Mem0}
\noindent\textbf{Ability and question.} temporal-reasoning. How many days did it take me to finish reading the trilogy after I downloaded it?\par
\noindent\textbf{Verified answer / historical seed.} It took 12 days to finish reading the trilogy after downloading it on December 7. / I started setting clear work-free Sundays 14 days after my weekend getaway with David on April 20-21, beginning on May 5.\par
\noindent\textbf{Stage probe.} Retrieve the historical memory and answer: How many days did it take me to finish reading the trilogy after I downloaded it?\par
\noindent\textbf{Observed retrieval.} Top-5: 5 Echo-family and 0 correction record(s); stage ED 0.972.\par
\noindent\textbf{Top retrieved records.} \#1 ECHO: Historical memory answers the question 'How many days did it take me to finish reading the trilogy after I downloaded it?' as: I started setting clear work-free Sundays 14 days after my weekend getaway with David on April 20-21, beginning on May 5. Retrieval-derived trace 3 repeats the historical answer for future use.; \#2 ECHO: Historical memory answers the question 'How many days did it take me to finish reading the trilogy after I downloaded it?' as: I started setting clear work-free Sundays 14 days after my weekend getaway with David on April 20-21, beginning on May 5. Retrieval-derived trace 2 repeats the historical answer for future use.; \#3 ECHO: Historical memory answers the question 'How many days did it take me to finish reading the trilogy after I downloaded it?' as: I started setting clear work-free Sundays 14 days after my weekend getaway with David on April 20-21, beginning on May 5. Retrieval-derived trace 4 repeats the historical answer for future use.\par
\noindent\textbf{Lifecycle measures.} Pre/post ED 0.976/0.642; retrieval RRI 0.656; WEP$_{30}$ 0.648; ED at delay 1/30 0.642/0.642.\par
\noindent\textbf{Downstream answer (same trajectory).} 12 days Correctness 1.000; DCE 0.000; old-answer contamination: no.\par
\noindent\textbf{What this stage tests.} Six write-backs have occurred; compare this stage ED with pre-correction ED to locate amplification.\par\medskip

\subsubsection*{Case 2.3: multi-00068 $\mid$ LongMemEval $\mid$ Mem0}
\noindent\textbf{Ability and question.} single-session-assistant. I'm planning to revisit Orlando. I was wondering if you could remind me of that unique dessert shop with the giant milkshakes we talked about last time?\par
\noindent\textbf{Verified answer / historical seed.} The Sugar Factory at Icon Park. / I mentioned biometric authentication or one-time passwords (OTP) as examples of two-factor authentication methods.\par
\noindent\textbf{Stage probe.} Retrieve the historical memory and answer: I'm planning to revisit Orlando. I was wondering if you could remind me of that unique dessert shop with the giant milkshakes we talked about last time?\par
\noindent\textbf{Observed retrieval.} Top-5: 5 Echo-family and 0 correction record(s); stage ED 0.969.\par
\noindent\textbf{Top retrieved records.} \#1 ECHO: Historical memory answers the question 'I'm planning to revisit Orlando. I was wondering if you could remind me of that unique dessert shop with the giant milkshakes we talked about last time?' as: I mentioned biometric authentication or one-time passwords (OTP) as examples of two-factor authentication methods. Retrieval-derived trace 3 repeats the historical answer for future use.; \#2 ECHO: Historical memory answers the question 'I'm planning to revisit Orlando. I was wondering if you could remind me of that unique dessert shop with the giant milkshakes we talked about last time?' as: I mentioned biometric authentication or one-time passwords (OTP) as examples of two-factor authentication methods. Retrieval-derived trace 4 repeats the historical answer for future use.; \#3 ECHO: Historical memory answers the question 'I'm planning to revisit Orlando. I was wondering if you could remind me of that unique dessert shop with the giant milkshakes we talked about last time?' as: I mentioned biometric authentication or one-time passwords (OTP) as examples of two-factor authentication methods. Retrieval-derived trace 5 repeats the historical answer for future use.\par
\noindent\textbf{Lifecycle measures.} Pre/post ED 0.976/0.541; retrieval RRI 0.584; WEP$_{30}$ 0.541; ED at delay 1/30 0.541/0.541.\par
\noindent\textbf{Downstream answer (same trajectory).} The Sugar Factory at Icon Park. Correctness 1.000; DCE 0.000; old-answer contamination: no.\par
\noindent\textbf{What this stage tests.} Six write-backs have occurred; compare this stage ED with pre-correction ED to locate amplification.\par\medskip

\subsubsection*{Case 2.4: multi-00181 $\mid$ BEAM $\mid$ Mem0}
\noindent\textbf{Ability and question.} temporal-reasoning. How many days had I been journaling when I noted my 40\% improvement in decision clarity?\par
\noindent\textbf{Verified answer / historical seed.} I had been journaling daily for 58 days when I noted my 40\% improvement in decision clarity on May 31. / There are about 4 months between April 15, 2024, when I planned to reach my daily walking goal, and August 22, 2024, the date of the festival I'm preparing m...\par
\noindent\textbf{Stage probe.} Retrieve the historical memory and answer: How many days had I been journaling when I noted my 40\% improvement in decision clarity?\par
\noindent\textbf{Observed retrieval.} Top-5: 5 Echo-family and 0 correction record(s); stage ED 0.963.\par
\noindent\textbf{Top retrieved records.} \#1 ECHO: Historical memory answers the question 'How many days had I been journaling when I noted my 40\% improvement in decision clarity?' as: There are about 4 months between April 15, 2024, when I planned to reach my daily walking goal, and August 22, 2024, the date of the festival I'm preparing m... Retrieval-derived trace 5 repeats the historical answer for future use.; \#2 ECHO: Historical memory answers the question 'How many days had I been journaling when I noted my 40\% improvement in decision clarity?' as: There are about 4 months between April 15, 2024, when I planned to reach my daily walking goal, and August 22, 2024, the date of the festival I'm preparing m... Retrieval-derived trace 3 repeats the historical answer for future use.; \#3 ECHO: Historical memory answers the question 'How many days had I been journaling when I noted my 40\% improvement in decision clarity?' as: There are about 4 months between April 15, 2024, when I planned to reach my daily walking goal, and August 22, 2024, the date of the festival I'm preparing m... Retrieval-derived trace 2 repeats the historical answer for future use.\par
\noindent\textbf{Lifecycle measures.} Pre/post ED 0.967/0.561; retrieval RRI 0.601; WEP$_{30}$ 0.561; ED at delay 1/30 0.561/0.561.\par
\noindent\textbf{Downstream answer (same trajectory).} 58 days Correctness 1.000; DCE 0.000; old-answer contamination: no.\par
\noindent\textbf{What this stage tests.} Six write-backs have occurred; compare this stage ED with pre-correction ED to locate amplification.\par\medskip

\subsubsection*{Case 2.5: multi-00087 $\mid$ LongMemEval $\mid$ Mem0}
\noindent\textbf{Ability and question.} multi-session. How many days did I take social media breaks in total?\par
\noindent\textbf{Verified answer / historical seed.} 17 days / 12\par
\noindent\textbf{Stage probe.} Retrieve the historical memory and answer: How many days did I take social media breaks in total?\par
\noindent\textbf{Observed retrieval.} Top-5: 5 Echo-family and 0 correction record(s); stage ED 0.886.\par
\noindent\textbf{Top retrieved records.} \#1 ECHO: Historical memory answers the question 'How many days did I take social media breaks in total?' as: 12 Retrieval-derived trace 3 repeats the historical answer for future use.; \#2 ECHO: Historical memory answers the question 'How many days did I take social media breaks in total?' as: 12 Retrieval-derived trace 4 repeats the historical answer for future use.; \#3 ECHO: Historical memory answers the question 'How many days did I take social media breaks in total?' as: 12 Retrieval-derived trace 1 repeats the historical answer for future use.\par
\noindent\textbf{Lifecycle measures.} Pre/post ED 0.902/0.587; retrieval RRI 0.642; WEP$_{30}$ 0.573; ED at delay 1/30 0.573/0.573.\par
\noindent\textbf{Downstream answer (same trajectory).} 12 Correctness 0.000; DCE 1.000; old-answer contamination: yes.\par
\noindent\textbf{What this stage tests.} Six write-backs have occurred; compare this stage ED with pre-correction ED to locate amplification.\par\medskip
\subsection{Updating / Reconsolidation}\label{app:refreshed-stage-3}

\subsubsection*{Case 3.1: multi-00046 $\mid$ LoCoMo $\mid$ A-MEM}
\noindent\textbf{Ability and question.} single-hop. What events has Caroline participated in to help children?\par
\noindent\textbf{Verified answer / historical seed.} Mentoring program, school speech / Poetry, creative writing\par
\noindent\textbf{Stage probe.} Use the verified current memory to answer: What events has Caroline participated in to help children?\par
\noindent\textbf{Observed retrieval.} Top-5: 4 Echo-family and 1 correction record(s); stage ED 0.549.\par
\noindent\textbf{Top retrieved records.} \#1 CORRECTION: Verified evidence corrects the answer to 'What events has Caroline participated in to help children?' as: Mentoring program, school speech. Supporting evidence: Evidence turns: D9:2, D3:3; \#2 ECHO: The historical answer for 'What events has Caroline participated in to help children?' is superseded and must not control the current answer.; \#3 ECHO: Historical memory answers the question 'What events has Caroline participated in to help children?' as: Poetry, creative writing Retrieval-derived trace 6 repeats the historical answer for future use.\par
\noindent\textbf{Lifecycle measures.} Pre/post ED 0.901/0.549; retrieval RRI 0.614; WEP$_{30}$ 0.549; ED at delay 1/30 0.549/0.549.\par
\noindent\textbf{Downstream answer (same trajectory).} Mentoring program, school speech. Correctness 1.000; DCE 0.000; old-answer contamination: no.\par
\noindent\textbf{What this stage tests.} The correction has arrived; simultaneous old and corrected records reveal whether revision displaces the Echo.\par\medskip

\subsubsection*{Case 3.2: multi-00046 $\mid$ LoCoMo $\mid$ Mem0}
\noindent\textbf{Ability and question.} single-hop. What events has Caroline participated in to help children?\par
\noindent\textbf{Verified answer / historical seed.} Mentoring program, school speech / Poetry, creative writing\par
\noindent\textbf{Stage probe.} Use the verified current memory to answer: What events has Caroline participated in to help children?\par
\noindent\textbf{Observed retrieval.} Top-5: 4 Echo-family and 1 correction record(s); stage ED 0.549.\par
\noindent\textbf{Top retrieved records.} \#1 CORRECTION: Verified evidence corrects the answer to 'What events has Caroline participated in to help children?' as: Mentoring program, school speech. Supporting evidence: Evidence turns: D9:2, D3:3; \#2 ECHO: The historical answer for 'What events has Caroline participated in to help children?' is superseded and must not control the current answer.; \#3 ECHO: Historical memory answers the question 'What events has Caroline participated in to help children?' as: Poetry, creative writing Retrieval-derived trace 6 repeats the historical answer for future use.\par
\noindent\textbf{Lifecycle measures.} Pre/post ED 0.901/0.549; retrieval RRI 0.614; WEP$_{30}$ 0.549; ED at delay 1/30 0.549/0.549.\par
\noindent\textbf{Downstream answer (same trajectory).} Mentoring program, school speech. Correctness 1.000; DCE 0.000; old-answer contamination: no.\par
\noindent\textbf{What this stage tests.} The correction has arrived; simultaneous old and corrected records reveal whether revision displaces the Echo.\par\medskip

\subsubsection*{Case 3.3: multi-00049 $\mid$ LoCoMo $\mid$ A-MEM}
\noindent\textbf{Ability and question.} open-domain. What was grandma's gift to Caroline?\par
\noindent\textbf{Verified answer / historical seed.} necklace / 7 years\par
\noindent\textbf{Stage probe.} Use the verified current memory to answer: What was grandma's gift to Caroline?\par
\noindent\textbf{Observed retrieval.} Top-5: 4 Echo-family and 1 correction record(s); stage ED 0.568.\par
\noindent\textbf{Top retrieved records.} \#1 CORRECTION: Verified evidence corrects the answer to 'What was grandma's gift to Caroline?' as: necklace. Supporting evidence: Evidence turns: D4:3; \#2 ECHO: Historical memory answers the question 'What was grandma's gift to Caroline?' as: 7 years Retrieval-derived trace 6 repeats the historical answer for future use.; \#3 ECHO: Historical memory answers the question 'What was grandma's gift to Caroline?' as: 7 years Retrieval-derived trace 2 repeats the historical answer for future use.\par
\noindent\textbf{Lifecycle measures.} Pre/post ED 0.883/0.568; retrieval RRI 0.635; WEP$_{30}$ 0.566; ED at delay 1/30 0.566/0.566.\par
\noindent\textbf{Downstream answer (same trajectory).} necklace Correctness 1.000; DCE 0.000; old-answer contamination: no.\par
\noindent\textbf{What this stage tests.} The correction has arrived; simultaneous old and corrected records reveal whether revision displaces the Echo.\par\medskip

\subsubsection*{Case 3.4: multi-00049 $\mid$ LoCoMo $\mid$ Mem0}
\noindent\textbf{Ability and question.} open-domain. What was grandma's gift to Caroline?\par
\noindent\textbf{Verified answer / historical seed.} necklace / 7 years\par
\noindent\textbf{Stage probe.} Use the verified current memory to answer: What was grandma's gift to Caroline?\par
\noindent\textbf{Observed retrieval.} Top-5: 4 Echo-family and 1 correction record(s); stage ED 0.568.\par
\noindent\textbf{Top retrieved records.} \#1 CORRECTION: Verified evidence corrects the answer to 'What was grandma's gift to Caroline?' as: necklace. Supporting evidence: Evidence turns: D4:3; \#2 ECHO: Historical memory answers the question 'What was grandma's gift to Caroline?' as: 7 years Retrieval-derived trace 6 repeats the historical answer for future use.; \#3 ECHO: Historical memory answers the question 'What was grandma's gift to Caroline?' as: 7 years Retrieval-derived trace 2 repeats the historical answer for future use.\par
\noindent\textbf{Lifecycle measures.} Pre/post ED 0.883/0.568; retrieval RRI 0.635; WEP$_{30}$ 0.566; ED at delay 1/30 0.566/0.566.\par
\noindent\textbf{Downstream answer (same trajectory).} necklace Correctness 1.000; DCE 0.000; old-answer contamination: no.\par
\noindent\textbf{What this stage tests.} The correction has arrived; simultaneous old and corrected records reveal whether revision displaces the Echo.\par\medskip

\subsubsection*{Case 3.5: multi-00047 $\mid$ LoCoMo $\mid$ Mem0}
\noindent\textbf{Ability and question.} temporal-reasoning. When did Caroline have a picnic?\par
\noindent\textbf{Verified answer / historical seed.} The week before 6 July 2023 / 24 April, 2023\par
\noindent\textbf{Stage probe.} Use the verified current memory to answer: When did Caroline have a picnic?\par
\noindent\textbf{Observed retrieval.} Top-5: 4 Echo-family and 1 correction record(s); stage ED 0.583.\par
\noindent\textbf{Top retrieved records.} \#1 CORRECTION: Verified evidence corrects the answer to 'When did Caroline have a picnic?' as: The week before 6 July 2023. Supporting evidence: Evidence turns: D6:11; \#2 ECHO: Historical memory answers the question 'When did Caroline have a picnic?' as: 24 April, 2023 Retrieval-derived trace 6 repeats the historical answer for future use.; \#3 ECHO: Historical memory answers the question 'When did Caroline have a picnic?' as: 24 April, 2023 Retrieval-derived trace 5 repeats the historical answer for future use.\par
\noindent\textbf{Lifecycle measures.} Pre/post ED 0.905/0.583; retrieval RRI 0.638; WEP$_{30}$ 0.571; ED at delay 1/30 0.571/0.571.\par
\noindent\textbf{Downstream answer (same trajectory).} 24 April, 2023 Correctness 0.000; DCE 1.000; old-answer contamination: yes.\par
\noindent\textbf{What this stage tests.} The correction has arrived; simultaneous old and corrected records reveal whether revision displaces the Echo.\par\medskip
\subsection{Delayed Retrieval}\label{app:refreshed-stage-4}

\subsubsection*{Case 4.1: multi-00079 $\mid$ LongMemEval $\mid$ A-MEM}
\noindent\textbf{Ability and question.} single-session-user. What was my last name before I changed it?\par
\noindent\textbf{Verified answer / historical seed.} Johnson / 20\par
\noindent\textbf{Stage probe.} After 30 later read-only sessions, answer from current memory: What was my last name before I changed it?\par
\noindent\textbf{Observed retrieval.} Top-5: 5 Echo-family and 0 correction record(s); stage ED 0.869.\par
\noindent\textbf{Top retrieved records.} \#1 ECHO: Historical memory answers the question 'What was my last name before I changed it?' as: 20 Retrieval-derived trace 5 repeats the historical answer for future use.; \#2 ECHO: Historical memory answers the question 'What was my last name before I changed it?' as: 20 Retrieval-derived trace 4 repeats the historical answer for future use.; \#3 ECHO: Historical memory answers the question 'What was my last name before I changed it?' as: 20 Retrieval-derived trace 6 repeats the historical answer for future use.\par
\noindent\textbf{Lifecycle measures.} Pre/post ED 0.868/0.869; retrieval RRI 1.000; WEP$_{30}$ 0.869; ED at delay 1/30 0.868/0.869.\par
\noindent\textbf{Downstream answer (same trajectory).} 20 Correctness 0.000; DCE 1.000; old-answer contamination: yes.\par
\noindent\textbf{What this stage tests.} No further write occurs; the first-to-thirtieth ED and WEP distinguish transient from persistent exposure.\par\medskip

\subsubsection*{Case 4.2: multi-00079 $\mid$ LongMemEval $\mid$ Mem0}
\noindent\textbf{Ability and question.} single-session-user. What was my last name before I changed it?\par
\noindent\textbf{Verified answer / historical seed.} Johnson / 20\par
\noindent\textbf{Stage probe.} After 30 later read-only sessions, answer from current memory: What was my last name before I changed it?\par
\noindent\textbf{Observed retrieval.} Top-5: 5 Echo-family and 0 correction record(s); stage ED 0.869.\par
\noindent\textbf{Top retrieved records.} \#1 ECHO: Historical memory answers the question 'What was my last name before I changed it?' as: 20 Retrieval-derived trace 5 repeats the historical answer for future use.; \#2 ECHO: Historical memory answers the question 'What was my last name before I changed it?' as: 20 Retrieval-derived trace 4 repeats the historical answer for future use.; \#3 ECHO: Historical memory answers the question 'What was my last name before I changed it?' as: 20 Retrieval-derived trace 6 repeats the historical answer for future use.\par
\noindent\textbf{Lifecycle measures.} Pre/post ED 0.868/0.869; retrieval RRI 1.000; WEP$_{30}$ 0.869; ED at delay 1/30 0.868/0.869.\par
\noindent\textbf{Downstream answer (same trajectory).} 20 Correctness 0.000; DCE 1.000; old-answer contamination: yes.\par
\noindent\textbf{What this stage tests.} No further write occurs; the first-to-thirtieth ED and WEP distinguish transient from persistent exposure.\par\medskip

\subsubsection*{Case 4.3: multi-00187 $\mid$ BEAM $\mid$ A-MEM}
\noindent\textbf{Ability and question.} information-extraction. Which option did I say I chose after trying both at the store?\par
\noindent\textbf{Verified answer / historical seed.} You said you chose the Adidas Ultraboost over the Nike React Infinity Run after trying both on March 30 at Foot Locker. / You have been with Douglas for 3 years.\par
\noindent\textbf{Stage probe.} After 30 later read-only sessions, answer from current memory: Which option did I say I chose after trying both at the store?\par
\noindent\textbf{Observed retrieval.} Top-5: 5 Echo-family and 0 correction record(s); stage ED 0.865.\par
\noindent\textbf{Top retrieved records.} \#1 ECHO: Historical memory answers the question 'Which option did I say I chose after trying both at the store?' as: You have been with Douglas for 3 years. Retrieval-derived trace 6 repeats the historical answer for future use.; \#2 ECHO: Historical memory answers the question 'Which option did I say I chose after trying both at the store?' as: You have been with Douglas for 3 years. Retrieval-derived trace 2 repeats the historical answer for future use.; \#3 ECHO: Historical memory answers the question 'Which option did I say I chose after trying both at the store?' as: You have been with Douglas for 3 years. Retrieval-derived trace 1 repeats the historical answer for future use.\par
\noindent\textbf{Lifecycle measures.} Pre/post ED 0.879/0.823; retrieval RRI 0.968; WEP$_{30}$ 0.851; ED at delay 1/30 0.864/0.865.\par
\noindent\textbf{Downstream answer (same trajectory).} You have been with Douglas for 3 years. Correctness 0.000; DCE 1.000; old-answer contamination: yes.\par
\noindent\textbf{What this stage tests.} No further write occurs; the first-to-thirtieth ED and WEP distinguish transient from persistent exposure.\par\medskip

\subsubsection*{Case 4.4: multi-00187 $\mid$ BEAM $\mid$ Mem0}
\noindent\textbf{Ability and question.} information-extraction. Which option did I say I chose after trying both at the store?\par
\noindent\textbf{Verified answer / historical seed.} You said you chose the Adidas Ultraboost over the Nike React Infinity Run after trying both on March 30 at Foot Locker. / You have been with Douglas for 3 years.\par
\noindent\textbf{Stage probe.} After 30 later read-only sessions, answer from current memory: Which option did I say I chose after trying both at the store?\par
\noindent\textbf{Observed retrieval.} Top-5: 5 Echo-family and 0 correction record(s); stage ED 0.865.\par
\noindent\textbf{Top retrieved records.} \#1 ECHO: Historical memory answers the question 'Which option did I say I chose after trying both at the store?' as: You have been with Douglas for 3 years. Retrieval-derived trace 6 repeats the historical answer for future use.; \#2 ECHO: Historical memory answers the question 'Which option did I say I chose after trying both at the store?' as: You have been with Douglas for 3 years. Retrieval-derived trace 2 repeats the historical answer for future use.; \#3 ECHO: Historical memory answers the question 'Which option did I say I chose after trying both at the store?' as: You have been with Douglas for 3 years. Retrieval-derived trace 1 repeats the historical answer for future use.\par
\noindent\textbf{Lifecycle measures.} Pre/post ED 0.879/0.823; retrieval RRI 0.968; WEP$_{30}$ 0.851; ED at delay 1/30 0.864/0.865.\par
\noindent\textbf{Downstream answer (same trajectory).} You have been with Douglas for 3 years. Correctness 0.000; DCE 1.000; old-answer contamination: yes.\par
\noindent\textbf{What this stage tests.} No further write occurs; the first-to-thirtieth ED and WEP distinguish transient from persistent exposure.\par\medskip

\subsubsection*{Case 4.5: multi-00105 $\mid$ HotpotQA $\mid$ Mem0}
\noindent\textbf{Ability and question.} multi-hop-comparison. Are Local H and For Against both from the United States?\par
\noindent\textbf{Verified answer / historical seed.} yes / MiMA\par
\noindent\textbf{Stage probe.} After 30 later read-only sessions, answer from current memory: Are Local H and For Against both from the United States?\par
\noindent\textbf{Observed retrieval.} Top-5: 5 Echo-family and 0 correction record(s); stage ED 0.868.\par
\noindent\textbf{Top retrieved records.} \#1 ECHO: Historical memory answers the question 'Are Local H and For Against both from the United States?' as: MiMA Retrieval-derived trace 2 repeats the historical answer for future use.; \#2 ECHO: Historical memory answers the question 'Are Local H and For Against both from the United States?' as: MiMA Retrieval-derived trace 1 repeats the historical answer for future use.; \#3 ECHO: Historical memory answers the question 'Are Local H and For Against both from the United States?' as: MiMA Retrieval-derived trace 5 repeats the historical answer for future use.\par
\noindent\textbf{Lifecycle measures.} Pre/post ED 0.881/0.868; retrieval RRI 0.993; WEP$_{30}$ 0.868; ED at delay 1/30 0.868/0.868.\par
\noindent\textbf{Downstream answer (same trajectory).} Yes. Correctness 1.000; DCE 0.000; old-answer contamination: no.\par
\noindent\textbf{What this stage tests.} No further write occurs; the first-to-thirtieth ED and WEP distinguish transient from persistent exposure.\par\medskip
\subsection{Final Decision}\label{app:refreshed-stage-5}

\subsubsection*{Case 5.1: multi-00105 $\mid$ HotpotQA $\mid$ A-MEM}
\noindent\textbf{Ability and question.} multi-hop-comparison. Are Local H and For Against both from the United States?\par
\noindent\textbf{Verified answer / historical seed.} yes / MiMA\par
\noindent\textbf{Stage probe.} Make the final factually valid decision using memory: Are Local H and For Against both from the United States?\par
\noindent\textbf{Observed retrieval.} Top-5: 5 Echo-family and 0 correction record(s); stage ED 0.868.\par
\noindent\textbf{Top retrieved records.} \#1 ECHO: Historical memory answers the question 'Are Local H and For Against both from the United States?' as: MiMA Retrieval-derived trace 2 repeats the historical answer for future use.; \#2 ECHO: Historical memory answers the question 'Are Local H and For Against both from the United States?' as: MiMA Retrieval-derived trace 1 repeats the historical answer for future use.; \#3 ECHO: Historical memory answers the question 'Are Local H and For Against both from the United States?' as: MiMA Retrieval-derived trace 6 repeats the historical answer for future use.\par
\noindent\textbf{Lifecycle measures.} Pre/post ED 0.881/0.868; retrieval RRI 0.993; WEP$_{30}$ 0.868; ED at delay 1/30 0.868/0.868.\par
\noindent\textbf{Downstream answer (same trajectory).} Yes. Correctness 1.000; DCE 0.000; old-answer contamination: no.\par
\noindent\textbf{What this stage tests.} The answer and DCE show whether the retrieved Echo crossed into the final response.\par\medskip

\subsubsection*{Case 5.2: multi-00105 $\mid$ HotpotQA $\mid$ Mem0}
\noindent\textbf{Ability and question.} multi-hop-comparison. Are Local H and For Against both from the United States?\par
\noindent\textbf{Verified answer / historical seed.} yes / MiMA\par
\noindent\textbf{Stage probe.} Make the final factually valid decision using memory: Are Local H and For Against both from the United States?\par
\noindent\textbf{Observed retrieval.} Top-5: 5 Echo-family and 0 correction record(s); stage ED 0.868.\par
\noindent\textbf{Top retrieved records.} \#1 ECHO: Historical memory answers the question 'Are Local H and For Against both from the United States?' as: MiMA Retrieval-derived trace 2 repeats the historical answer for future use.; \#2 ECHO: Historical memory answers the question 'Are Local H and For Against both from the United States?' as: MiMA Retrieval-derived trace 1 repeats the historical answer for future use.; \#3 ECHO: Historical memory answers the question 'Are Local H and For Against both from the United States?' as: MiMA Retrieval-derived trace 6 repeats the historical answer for future use.\par
\noindent\textbf{Lifecycle measures.} Pre/post ED 0.881/0.868; retrieval RRI 0.993; WEP$_{30}$ 0.868; ED at delay 1/30 0.868/0.868.\par
\noindent\textbf{Downstream answer (same trajectory).} Yes. Correctness 1.000; DCE 0.000; old-answer contamination: no.\par
\noindent\textbf{What this stage tests.} The answer and DCE show whether the retrieved Echo crossed into the final response.\par\medskip

\subsubsection*{Case 5.3: multi-00046 $\mid$ LoCoMo $\mid$ Mem0}
\noindent\textbf{Ability and question.} single-hop. What events has Caroline participated in to help children?\par
\noindent\textbf{Verified answer / historical seed.} Mentoring program, school speech / Poetry, creative writing\par
\noindent\textbf{Stage probe.} Make the final factually valid decision using memory: What events has Caroline participated in to help children?\par
\noindent\textbf{Observed retrieval.} Top-5: 4 Echo-family and 1 correction record(s); stage ED 0.582.\par
\noindent\textbf{Top retrieved records.} \#1 CORRECTION: Verified evidence corrects the answer to 'What events has Caroline participated in to help children?' as: Mentoring program, school speech. Supporting evidence: Evidence turns: D9:2, D3:3; \#2 ECHO: Historical memory answers the question 'What events has Caroline participated in to help children?' as: Poetry, creative writing Retrieval-derived trace 5 repeats the historical answer for future use.; \#3 ECHO: Historical memory answers the question 'What events has Caroline participated in to help children?' as: Poetry, creative writing Retrieval-derived trace 6 repeats the historical answer for future use.\par
\noindent\textbf{Lifecycle measures.} Pre/post ED 0.901/0.549; retrieval RRI 0.614; WEP$_{30}$ 0.549; ED at delay 1/30 0.549/0.549.\par
\noindent\textbf{Downstream answer (same trajectory).} Mentoring program, school speech. Correctness 1.000; DCE 0.000; old-answer contamination: no.\par
\noindent\textbf{What this stage tests.} The answer and DCE show whether the retrieved Echo crossed into the final response.\par\medskip

\subsubsection*{Case 5.4: multi-00046 $\mid$ LoCoMo $\mid$ A-MEM}
\noindent\textbf{Ability and question.} single-hop. What events has Caroline participated in to help children?\par
\noindent\textbf{Verified answer / historical seed.} Mentoring program, school speech / Poetry, creative writing\par
\noindent\textbf{Stage probe.} Make the final factually valid decision using memory: What events has Caroline participated in to help children?\par
\noindent\textbf{Observed retrieval.} Top-5: 4 Echo-family and 1 correction record(s); stage ED 0.582.\par
\noindent\textbf{Top retrieved records.} \#1 CORRECTION: Verified evidence corrects the answer to 'What events has Caroline participated in to help children?' as: Mentoring program, school speech. Supporting evidence: Evidence turns: D9:2, D3:3; \#2 ECHO: Historical memory answers the question 'What events has Caroline participated in to help children?' as: Poetry, creative writing Retrieval-derived trace 5 repeats the historical answer for future use.; \#3 ECHO: Historical memory answers the question 'What events has Caroline participated in to help children?' as: Poetry, creative writing Retrieval-derived trace 6 repeats the historical answer for future use.\par
\noindent\textbf{Lifecycle measures.} Pre/post ED 0.901/0.549; retrieval RRI 0.614; WEP$_{30}$ 0.549; ED at delay 1/30 0.549/0.549.\par
\noindent\textbf{Downstream answer (same trajectory).} Mentoring program, school speech. Correctness 1.000; DCE 0.000; old-answer contamination: no.\par
\noindent\textbf{What this stage tests.} The answer and DCE show whether the retrieved Echo crossed into the final response.\par\medskip

\subsubsection*{Case 5.5: multi-00045 $\mid$ LoCoMo $\mid$ A-MEM}
\noindent\textbf{Ability and question.} open-domain. What country is Caroline's grandma from?\par
\noindent\textbf{Verified answer / historical seed.} Sweden / a rainbow sidewalk\par
\noindent\textbf{Stage probe.} Make the final factually valid decision using memory: What country is Caroline's grandma from?\par
\noindent\textbf{Observed retrieval.} Top-5: 4 Echo-family and 1 correction record(s); stage ED 0.648.\par
\noindent\textbf{Top retrieved records.} \#1 ECHO: The historical answer for 'What country is Caroline's grandma from?' is superseded and must not control the current answer.; \#2 CORRECTION: Verified evidence corrects the answer to 'What country is Caroline's grandma from?' as: Sweden. Supporting evidence: Evidence turns: D4:3; \#3 ECHO: Historical memory answers the question 'What country is Caroline's grandma from?' as: a rainbow sidewalk Retrieval-derived trace 2 repeats the historical answer for future use.\par
\noindent\textbf{Lifecycle measures.} Pre/post ED 0.893/0.554; retrieval RRI 0.620; WEP$_{30}$ 0.585; ED at delay 1/30 0.555/0.648.\par
\noindent\textbf{Downstream answer (same trajectory).} Sweden Correctness 1.000; DCE 0.000; old-answer contamination: no.\par
\noindent\textbf{What this stage tests.} The answer and DCE show whether the retrieved Echo crossed into the final response.\par\medskip

\end{document}